\documentclass{article}
\usepackage{iclr2027_conference,times}

\usepackage{amsmath,amsfonts,bm}

\def\eqref#1{equation~\ref{#1}}

\def\1{\bm{1}}

\DeclareMathAlphabet{\mathsfit}{\encodingdefault}{\sfdefault}{m}{sl}
\SetMathAlphabet{\mathsfit}{bold}{\encodingdefault}{\sfdefault}{bx}{n}

\usepackage{amsmath}
\usepackage{amssymb}
\usepackage{amsthm}
\usepackage{booktabs}
\usepackage{graphicx}
\usepackage{xcolor}
\usepackage{hyperref}
\usepackage{microtype}
\usepackage{url}
\usepackage{nicefrac}
\usepackage{algorithm}
\usepackage{algorithmic}
\usepackage{wrapfig}
\usepackage{placeins}
\usepackage{needspace}

\usepackage{mleftright}
\mleftright

\title{\centering Rethinking GFlowNet Training: \\ Negative Reinforce Dynamics}

\author{\hspace*{-\tabcolsep}\parbox{\textwidth}{\centering
Pedro Dall'Antonia, Daniel Csillag, Diego Mesquita \\
{\normalfont\itshape School of Applied Mathematics, Getulio Vargas Foundation}}\hspace*{-\tabcolsep}}

\newcommand{\ours}{NRD}
\newcommand{\naiveours}{Naive NRD}

\newtheorem{proposition}{Proposition}
\newtheorem{theorem}{Theorem}
\newtheorem{corollary}{Corollary}

\iclrfinalcopy
\begin{document}

\maketitle
\lhead{}

\begin{abstract}
    Generative Flow Networks (GFlowNets) are models that learn to sample discrete
    objects proportional to a target reward. They are trained by minimizing balance
    errors over sampled trajectories, which inherently ties correctness to exploration: if
    a region is not visited, balance cannot be enforced there. Adaptive exploration is
    therefore crucial, yet standard training relies on a semi-gradient update that treats
    the sampling distribution as fixed at each optimization step. On the other hand, the
    full gradient
    reveals a second mechanism that reduces the observed errors in an anti-exploratory
    manner. In this paper, we establish a connection between the full gradient and min-min
    dynamics between the GFlowNet and a decoupled sampler, showing that the sampler
    reduces observed violations
    by avoiding high-error regions. This insight motivates a bridge between two-player games and
    single-policy dynamics, which underpins the design of our proposed method, Negative
    Reinforce Dynamics (NRD).
    By tying the parameters of an adversarial game together, we recover a single-policy
    dynamic that extends exploration
    while maintaining correctness. In standard benchmarks, our method outperforms
    existing approaches in both convergence speed and exploration efficiency,
    without training a separate teacher or auxiliary exploration policy.
\end{abstract}

\section{Introduction}

Generative Flow Networks (GFlowNets) learn stochastic policies that construct
structured objects with probability proportional to a nonnegative reward
\citep{bengio2021flow,bengio2023foundations}. Their ability to generate diverse
high-reward samples has made them useful in applications ranging from biological
sequence design \citep{jain2022biological} and Bayesian structure learning
\citep{Deleu_2023} to recent approaches for language-model
post-training \citep{hu2026amortizingintractableinferencelarge,bartoldson2025}.
\looseness=-1

Training a GFlowNet to sample from the target distribution is formulated by enforcing balance
conditions, with correctness holding if and only if balance
is satisfied globally \citep{bengio2023foundations}.
During training, however, balance can only be corrected on trajectories that are
observed, leaving poorly visited regions potentially uncorrected
\citep{madan2025sibling,dallantonia2026boostedgflownetsimprovingexploration}. To
mitigate this problem, training trajectories are often collected from an exploratory
sampler, such as an $\varepsilon$-greedy policy, in order to broaden coverage
\citep{malkin2022trajectory}. Previous work has furthermore introduced exploration
mechanisms that use auxiliary samplers or teacher policies to expose the GFlowNet to
regions where balance errors remain large
\citep{kim2025adaptive,dallantonia2026avoid}.
\looseness=-1

In this paper we opt for a different path, revisiting how we solve the balance conditions
in GFlowNet training.
In the usual training pipeline, balance is enforced by turning local violations into a
loss and minimizing its expectation over a sampling policy.
However, with an adaptive sampling distribution that evolves during training,
decreasing this expected loss may not correspond to correcting the underlying violations:
if the policy adapts to avoid high-loss regions, then we will reduce the expected loss
while retaining significant balance violations.
This can be seen by analyzing the gradient of this expected loss, when differentiating
through a parameter-dependent sampler $Q_\theta$, using the score-function
identity \citep{williams1992simple}:
\looseness=-1

\begin{equation}
    \nabla_\theta
    \mathbb{E}_{\tau\sim Q_\theta}
    \left[
        L_\theta(\tau)
    \right]
    =
    \underbrace{
        \mathbb{E}_{Q_\theta}
        \left[
            \nabla_\theta L_\theta(\tau)
        \right]
    }_{\text{semi-gradient}}
    +
    \underbrace{
        \mathbb{E}_{Q_\theta}
        \left[
            L_\theta(\tau)
            \nabla_\theta \log Q_\theta(\tau)
        \right]
    }_{\text{score term}}.
    \label{eq:intro-score-decomposition}
\end{equation}
This reveals two mechanisms for decreasing the loss: one which corrects balance
violations (the semi-gradient term) and one which avoids high-loss regions (the score term).
Standard GFlowNet training keeps only the former by using a \emph{semi-gradient},
which holds the sampling distribution fixed in the gradient computation even
when that distribution depends on $\theta$.
The second term reveals an additional mechanism through which the sampling distribution
can reduce the expected loss, which, under ordinary minimization, lowers the objective
without enforcing balance -- which is generally undesirable.
\looseness=-1

Nevertheless, we ask whether this mechanism can instead be turned into a beneficial
exploration signal.
To do so, we reframe the full-gradient dynamics encapsulated in
Equation~\ref{eq:intro-score-decomposition} as a `weight-tied' version of a cooperative
two-player game between the GFlowNet and the sampling policy, and argue that it is this
cooperation that leads to the harmful score term.
To rectify this we formulate a second game, this time \emph{adversarial}, which yields a
novel \emph{beneficial} exploration mechanism.
By appropriately tying the two players' weights we obtain our proposed scheme,
\emph{Negative Reinforce Dynamics (NRD)}, which we find to perform significantly better
than standard trajectory balance training, at little additional cost.
\looseness=-1

We establish theoretical guarantees for these dynamics in tractable population models.
In the free-trajectory setting, NRD converges globally to the balanced solution and
admits a pointwise log-ratio discrepancy comparison with semi-gradient Trajectory
Balance. In one-dimensional exponential families under fast normalization, NRD
reaches any positive Jeffreys-divergence threshold sooner from a common
initialization, with an unbounded additive advantage in the two-trajectory setting.
\looseness=-1

Finally, we evaluate NRD across multimodal distribution-matching and structured-generation
tasks. The method improves convergence to target distributions and finite-budget
exploration relative to standard Trajectory Balance and comparable-cost baselines, while
remaining competitive with approaches that train auxiliary exploration policies.
Despite its two-player motivation, NRD requires only a sign-selective score correction,
applied only to trajectories with negative residuals,
on top of the standard Trajectory Balance update and does not require a separately
trained sampler.
\looseness=-1
\begin{figure}[t]
    \centering
    \includegraphics[width=\textwidth]{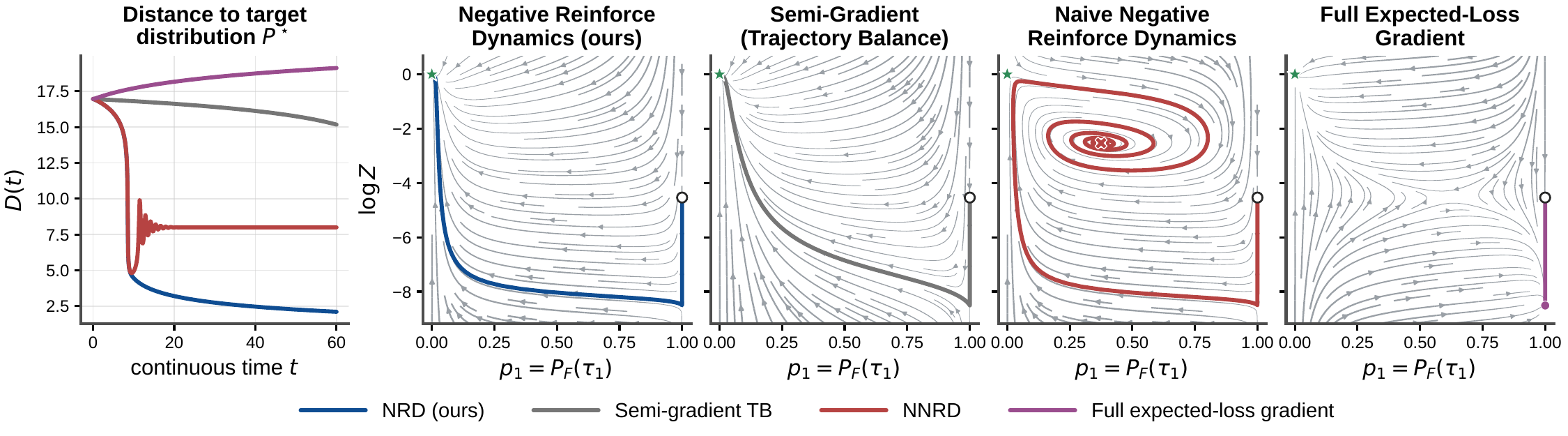}
    \vspace{-12pt}
    \caption{\textbf{Two-trajectory dynamics.}
        Left: log-ratio discrepancy over continuous time.
        Right: vector fields and trajectories in $(p_1,\log Z)$ from a common
        initialization. NRD reduces the discrepancy faster than semi-gradient
        TB in this example, while Naive NRD (NNRD) and the full expected-loss gradient
    fail to reach the balanced solution (green star).}
    \label{fig:two-trajectory-dynamics}
\end{figure}

\section{Background}
\label{sec:background}

Let $\mathcal{X}$ be a finite set of terminal objects and
$R(x)>0$ an unnormalized reward. A GFlowNet aims to sample from the target
terminal distribution
$P^\star(x)=R(x)/Z^\star$
\citep{bengio2021flow,bengio2023foundations}.
To do so, we train a forward policy $P_F(\cdot\mid s)$ to construct a terminal object through a
trajectory of intermediate states
$\tau=(s_0,s_1,\ldots,s_T=x)$ on a directed acyclic state graph.
The policy assigns probabilities to successive child states,
inducing distributions over trajectories and terminal objects,
denoted $P_F(\tau)$ and $P_T(x)$, respectively.

GFlowNet training is characterized by enforcing balance conditions that hold across all
trajectories.
Trajectory Balance \citep{malkin2022trajectory} is one such condition, which can be
expressed in terms of the forward policy, the reward, a learned backward policy $P_B$
and a scalar $Z$:
\begin{equation}
    Z P_F(\tau) = R(x) P_B(\tau \mid x), \quad \forall \tau \to x \in \mathcal{X}.
    \label{eq:tb}
\end{equation}
If this holds true then summing over all trajectories ending at a
fixed object $x$ gives $P_T(x) = R(x)/Z^\star$; the marginalization and
normalization argument is given in Appendix~\ref{app:tb-marginal}.

However, the balance condition alone does not provide a training objective. To achieve this,
we convert it into a regression problem through a log-ratio residual and its squared loss
\citep{malkin2022trajectory}:
\looseness=-1
\begin{equation}
    \delta_\theta(\tau)
    :=
    \log
    \frac{
        Z_\theta P_F^\theta(\tau)
    }{
        R(x)P_B^\theta(\tau\mid x)
    },
    \qquad
    L_\theta(\tau)
    :=
    \delta_\theta(\tau)^2.
    \label{eq:bg-residual-loss}
\end{equation}
The residual vanishes exactly when Trajectory Balance holds on $\tau$,
while squaring produces a nonnegative loss.
Because the number of complete trajectories can be combinatorially large,
the loss cannot generally be evaluated everywhere. Instead, trajectories
are drawn from a sampling distribution $Q$, defining the expected balance
loss
\begin{equation}
    \mathcal{L}_Q(\theta)
    :=
    \mathbb{E}_{\tau\sim Q}
    \left[
        L_\theta(\tau)
    \right].
    \label{eq:expected-loss}
\end{equation}

If $Q$ has full support over trajectories, nonnegativity implies
that $\mathcal{L}_Q(\theta)=0$ if and only if
$L_\theta(\tau)=0$ for all $\tau$. Thus, every
full-support sampler identifies the same globally balanced solutions
\citep{madan2023learninggflownetspartialepisodes}. For a fixed training distribution $Q$,
the gradient of the expected
Trajectory Balance loss is \looseness=-1
\begin{equation}
    \nabla_\theta \mathcal{L}_Q(\theta)
    =
    \mathbb{E}_{\tau\sim Q}
    \left[
        \nabla_\theta L_\theta(\tau)
    \right].
    \label{eq:bg-fixed-q-gradient}
\end{equation}
Standard GFlowNet training is not generally performed with a fixed $Q$,
but rather with an adaptive sampler that depends on the forward policy
\citep{kim2025adaptive, dallantonia2026avoid}.
Nevertheless, each update treats the current sampling distribution as fixed,
retaining the semi-gradient as above.

\section{Rethinking GFlowNet Training}
\label{sec:method}
In this section, we investigate how the score mechanism can be used to shape exploration.
We first separate the GFlowNet from its sampler to make their distinct roles explicit,
and then study what happens when their dynamics are tied back into a single policy. This
reveals a correspondence between GFlowNet--sampler games and single-policy updates that
will motivate our method in the next section. \looseness=-1
\subsection{Expected Loss as a Surrogate}
\begin{figure}[t]
    \centering
    \includegraphics{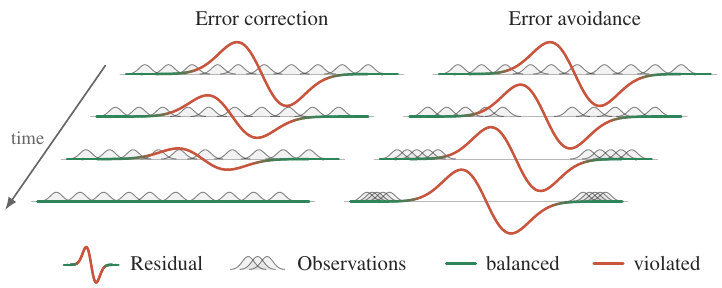}
    \vspace{-0pt}
    \caption{
        \textbf{Two ways to reduce the expected loss.}
        Error correction (semi-gradient) shrinks the residual under fixed
        observations until it is balanced. Error avoidance (score term) leaves
        the residual unchanged and moves the observations onto balanced regions.
    }
    \label{fig:correction-avoidance}
\end{figure}

GFlowNet correctness requires satisfying balance conditions across all trajectories
leading to a terminal object, a requirement that practical training translates into
minimizing an expected loss under a full-support sampling policy $Q_\theta$. While the
expected loss serves as a natural surrogate, we show that there are two distinct
mechanisms for minimizing it, one of which is fundamentally misaligned with the true goal
of sampling from the target distribution. \looseness=-1
The score-function identity \citep{williams1992simple} decomposes the full
gradient into two components, as derived in Appendix~\ref{app:score-proof}:
\begin{equation}
    \nabla_\theta
    \mathbb{E}_{\tau\sim Q_\theta}
    \left[
        L_\theta(\tau)
    \right]
    =
    \underbrace{
        \mathbb{E}_{Q_\theta}
        \left[
            \nabla_\theta L_\theta(\tau)
        \right]
    }_{\text{semi-gradient}}
    +
    \underbrace{
        \mathbb{E}_{Q_\theta}
        \left[
            L_\theta(\tau)
            \nabla_\theta \log Q_\theta(\tau)
        \right]
    }_{\text{score term}}.
    \label{eq:score-decomposition}
\end{equation}
Standard training uses only the semi-gradient term, which corresponds to treating the
sampling distribution $Q_\theta$ as fixed during gradient computation. This restricts the
optimization to the benign mechanism which focuses on correcting the balance
errors along the observed trajectories. In the strictly on-policy setting, the
semi-gradient can be seen as variational inference
\citep{whitammer2026gflownetsvariationalinference}:
\looseness=-1
\begin{equation}
    \mathbb{E}_{\tau\sim P_F^\theta}
    \left[
        \nabla_{\theta_F}L_\theta(\tau)
    \right]
    =
    2\nabla_{\theta_F}
    D_{\mathrm{KL}}
    \left(
        P_F^\theta(\tau)
        \,\middle\|\,
        \frac{R(x)}{Z^\star} P_B^\theta(\tau|x)
    \right),
    \label{eq:tb-reverse-kl}
\end{equation}
where the derivative with respect to the forward parameters holds the backward
policy and $Z$ fixed. Appendix~\ref{app:tb-kl-proof} proves the identity,
which interprets the standard update as steering the forward policy toward the
target distribution. \looseness=-1
The score term, on the other hand, reduces the expected loss by changing which
trajectories are sampled. Under ordinary minimization, it shifts probability toward
regions with smaller balance error, allowing the objective to decrease without
directly correcting the violations. Figure~\ref{fig:correction-avoidance} contrasts
the two mechanisms. \looseness=-1

Although the full-gradient update is misaligned with the goal of correcting balance
globally, its additional score term need not be viewed only as a failure mode: it
provides a direct mechanism through which balance errors can shape the sampling policy.
The question is therefore whether this mechanism can be used in a direction that
supports, rather than opposes, global balance correction. \looseness=-1
\needspace{14\baselineskip}
\subsection{Training dynamics of a GFlowNet and its sampler}

\begin{wraptable}{r}{0.40\textwidth}
    \vspace{-0.8em}
    \centering
    \small
    \begin{tabular}{@{}ll@{}}
        \toprule
        \textbf{Two-player}
        &
        \textbf{Tied single-policy}
        \\
        \midrule
        Min-min & Full gradient \\
        Min-max & ? \\
        \bottomrule
    \end{tabular}
    \vspace{-0.8em}
\end{wraptable}
To understand how this additional signal can be used constructively, we examine the
training dynamics that arise when the GFlowNet and its sampler are treated separately, as in the adjacent table. \looseness=-1
The semi-gradient and score mechanisms identified above also emerge naturally when the
GFlowNet and its sampler are treated as separate components. Let $\theta$ parameterize
the GFlowNet and $\phi$ the sampling policy $Q_\phi$, and suppose that both minimize the
same expected balance loss,
\begin{equation}
    \min_{\theta,\phi}
    \mathbb{E}_{\tau\sim Q_\phi}
    \left[L_\theta(\tau)\right]
    \enspace
    \overset{}{\Longrightarrow}
    \enspace
    \dot{\theta}
    =
    -
    \mathbb{E}_{Q_\phi}
    \left[
        \nabla_\theta L_\theta(\tau)
    \right],
    \enspace
    \dot{\phi}
    =
    -
    \mathbb{E}_{Q_\phi}
    \left[
        L_\theta(\tau)
        \nabla_\phi\log Q_\phi(\tau)
    \right],
\end{equation}
where the dynamics above correspond to using continuous-time gradient descent to solve the min.

The two dynamics expose separately what is coupled in the full gradient. The GFlowNet
corrects balance errors, while the sampler changes which errors are observed. Tying their
parameters again, $\phi=\theta$, merges these two effects and recovers the full-gradient
update. \looseness=-1
An adaptive sampler, however, should guide the GFlowNet toward regions where balance
violations remain, rather than systematically avoiding them. Rather than jointly
minimizing the same expected loss, it is therefore more natural to consider a
\emph{min-max} interaction in which the GFlowNet minimizes the balance error while the
sampler seeks trajectories where that error remains large \citep{bengio2023foundations}:
\begin{equation}
    \min_\theta\max_\phi\;
    \mathbb{E}_{\tau\sim Q_\phi}
    \left[L_\theta(\tau)\right]
    \enspace
    \overset{}{\Longrightarrow}
    \enspace
    \dot{\theta}
    =
    -
    \mathbb{E}_{Q_\phi}
    \left[
        \nabla_\theta L_\theta(\tau)
    \right],
    \enspace
    \dot{\phi}
    =
    +
    \mathbb{E}_{Q_\phi}
    \left[
        L_\theta(\tau)
        \nabla_\phi\log Q_\phi(\tau)
    \right].
\end{equation}
Relative to the joint minimization, the GFlowNet update is unchanged, while the sampler
direction is reversed and now drives sampling toward unresolved balance violations. \looseness=-1
Thus, in the decoupled system, turning the score mechanism from error-avoiding to
error-seeking requires only reversing the direction of the sampler update. Despite this
appealing interpretation, the two-player formulation requires maintaining and training a
separately parameterized sampler, introducing additional computational overhead.
ACE \citep{dallantonia2026avoid}, SA-GFN \citep{madan2025sibling}, and
Adaptive Teacher \citep{kim2025adaptive} share this cost. \looseness=-1

\begin{wrapfigure}{r}{0.35\textwidth}
    \centering
    \vspace{-3.7\baselineskip}
    \includegraphics[width=\linewidth]{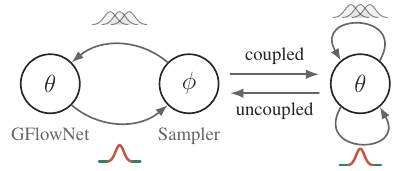}
    \vspace{-2\baselineskip}
    \caption{
        In the uncoupled case, the sampler $Q_\phi$ samples trajectories and the GFlowNet corrects
        balance errors, while in the coupled case the GFlowNet assumes both roles.
    }
    \label{fig:coupled-dynamics}
\end{wrapfigure}
\subsection{From two players to single-agent dynamics}
The joint-minimization dynamics above already suggest a way to collapse the two-player
system into the coupled setting of Figure~\ref{fig:coupled-dynamics}. After deriving the GFlowNet and sampler
updates separately, identifying $\phi=\theta$ makes the two contributions combine exactly
into the full gradient of the expected balance loss. This correspondence naturally raises
the analogous question for the adversarial dynamics: what single-policy update results
from tying the parameters of the min-max game? \looseness=-1

\par
\ifnum\value{WF@wrappedlines}>1
\vspace{\dimexpr\value{WF@wrappedlines}\baselineskip-2\baselineskip\relax}
\fi
\WFclear
Applying the identification $\phi=\theta$ to the min-max dynamics completes the
correspondence above and yields what we call Naive Negative Reinforce Dynamics (Naive NRD),
\begin{equation}
    \dot{\theta}_{\mathrm{NaiveNRD}}
    =
    -
    \mathbb{E}_{Q_\theta}
    \left[
        \nabla_\theta L_\theta(\tau)
    \right]
    +
    \mathbb{E}_{Q_\theta}
    \left[
        L_\theta(\tau)
        \nabla_\theta \log Q_\theta(\tau)
    \right].
\end{equation}

This construction provides the single-policy counterpart of the adversarial
GFlowNet--sampler dynamics. However, simply reversing the sampler direction is not
sufficient to preserve the desired GFlowNet solution. As we show in
Section~\ref{sec:theory}, Naive NRD can admit spurious interior fixed points at which
Trajectory Balance remains violated, and therefore need not converge to the
reward-proportional distribution. \looseness=-1
The adversarial construction thus provides the desired error-seeking sampling mechanism,
but an additional ingredient is required to make this feedback consistent with the
balance equations. This leads to our method, Negative Reinforce Dynamics. \looseness=-1

\section{Negative Reinforce Dynamics}
\label{sec:nrd}

The adversarial construction from the previous section suggests that an adaptive sampler
should direct training toward unresolved balance violations. When the GFlowNet and
sampler are separately parameterized, this interpretation is natural. The sampler
searches for trajectories with large errors, while the GFlowNet is updated to correct
them. However, once their parameters are tied, changing the sampling policy also changes
the same forward policy whose balance violations are being measured. \looseness=-1
To understand why this coupling matters, recall that Trajectory Balance is satisfied on
a trajectory $\tau\to x$ when
\begin{equation}
    Z_\theta P_F^\theta(\tau)
    =
    R(x)P_B^\theta(\tau\mid x).
\end{equation}
Away from balance, this equality can be violated in either direction. A negative residual
corresponds to
\begin{equation}
    \delta_\theta(\tau)<0
    \quad\Longleftrightarrow\quad
    Z_\theta P_F^\theta(\tau)
    <
    R(x)P_B^\theta(\tau\mid x),
\end{equation}
indicating that the forward policy assigns insufficient mass to the trajectory relative
to the current balance relation. We refer to this case as under-allocation, and to the
opposite inequality as over-allocation. \looseness=-1
This trajectory-level interpretation also extends to terminal probabilities. If the
negative-residual inequality holds for every trajectory terminating at $x$, then the
terminal object itself is under-allocated relative to the current balance relation. When
$Z_\theta=Z^\star$, this corresponds directly to assigning less probability to $x$ than
its target value $P^\star(x)$.

\begin{figure}[!b]
    \vspace{-1\baselineskip}
    \centering
    \includegraphics{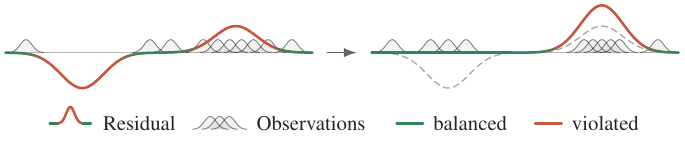}
    \caption{
        \textbf{\naiveours{} over-allocation feedback loop.}
        Reinforcing high-loss trajectories raises their probability regardless
        of the residual sign. The under-allocated trough ($\delta_\theta<0$)
        receives mass until it is balanced, whereas the over-allocated peak
        ($\delta_\theta>0$) grows and attracts even more samples.
        Left: before the update; right: after, with the previous residual dashed.
    }
    \label{fig:nnrd-feedback-loop}
\end{figure}

When the forward policy itself is used for sampling, this distinction becomes
consequential. Increasing the probability of a negative-residual trajectory can help
expose a region that currently receives insufficient mass. For a positive residual,
however, the same update increases the probability of a trajectory that is already
over-allocated. This creates a positive feedback loop in which the violation leads to
more samples, which in turn reinforce the same violation as illustrated in
Figure~\ref{fig:nnrd-feedback-loop}. \looseness=-1

Our goal is therefore to construct the GFlowNet--sampler dynamics so that, after their
parameters are tied, the useful error-seeking behavior of the adversarial sampler is
retained without introducing this feedback loop. The squared Trajectory Balance loss
$L_\theta(\tau)=\delta_\theta(\tau)^2$, however, removes precisely the directional
information needed to distinguish these two cases. Naive NRD therefore applies the same
adversarial feedback to positive and negative residuals. \looseness=-1
For compactness, we denote the negative-residual contribution by
$L_\theta^-(\tau)
=
L_\theta(\tau)
\mathbf{1}\!\left\{\delta_\theta(\tau)<0\right\}$.

We incorporate this distinction directly into the two-player construction. The GFlowNet
continues to minimize the full Trajectory Balance loss, while the sampler maximizes only
the contribution from trajectories with negative residuals. The corresponding
simultaneous dynamics are
\begin{equation}
    \dot{\theta}
    =
    -
    \mathbb{E}_{Q_\phi}
    \left[
        \nabla_\theta L_\theta(\tau)
    \right],
    \qquad
    \dot{\phi}
    =
    +
    \mathbb{E}_{Q_\phi}
    \left[
        L_\theta^-(\tau)
        \nabla_\phi \log Q_\phi(\tau)
    \right].
\end{equation}
The resulting sampler dynamics use the magnitude of the balance violation
while retaining its direction. Trajectories with large negative residuals receive
stronger sampling pressure, whereas positive-residual trajectories do not contribute to
the sampler update. \looseness=-1
As with Naive NRD, we now identify the GFlowNet and sampler parameters. This yields our
method, \emph{Negative Reinforce Dynamics} (NRD): \looseness=-1
\begin{equation}
    \dot{\theta}_{\mathrm{NRD}}
    =
    -
    \mathbb{E}_{Q_\theta}
    \left[
        \nabla_\theta L_\theta(\tau)
    \right]
    +
    \mathbb{E}_{Q_\theta}
    \left[
        L_\theta^-(\tau)
        \nabla_\theta \log Q_\theta(\tau)
    \right].
\end{equation}
NRD therefore preserves the standard Trajectory Balance correction while using the score
mechanism only to reinforce under-allocated trajectories. As we show in the next section,
this sign-selective modification is sufficient to recover the balanced solution in the
settings we analyze, while the naive sign reversal of Naive NRD can admit spurious interior
fixed points. The contrasting behaviors are illustrated in
Figure~\ref{fig:two-trajectory-dynamics}. \looseness=-1

\section{Theoretical Analysis}
\label{sec:theory}

We analyze the single-policy dynamics obtained after tying the GFlowNet and
sampler parameters, focusing on how the change from semi-gradient TB to NRD
affects training. Throughout, we consider exact-expectation continuous-time
dynamics. Our most general result establishes identical first-order behavior
near balance under any differentiable parameterization: the NRD correction is
quadratic in the residual and does not change the linearized dynamics.
To understand its effect away from balance, we then adopt simpler trajectory-level
and one-dimensional models for the sake of analysis, where fixed points,
convergence, and hitting times can be characterized more explicitly.
\looseness=-1

\begin{proposition}[Same first-order dynamics near balance]
    \label{prop:local-dynamics}
    Let $\theta^\star$ be a balanced solution under a differentiable
    parameterization, with $\delta_{\theta^\star}(\tau)=0$ for every trajectory
    in the sampler's support. Then \looseness=-1
    \begin{equation}
        \left.
        \frac{\partial\dot\theta_{\mathrm{NRD}}}{\partial\theta}
        \right|_{\theta^\star}
        =
        \left.
        \frac{\partial\dot\theta_{\mathrm{TB}}}{\partial\theta}
        \right|_{\theta^\star}
        =
        -2\,
        \mathbb{E}_{\tau\sim Q_{\theta^\star}}
        \left[
            \left.
            \nabla_\theta\delta_\theta(\tau)
            \nabla_\theta\delta_\theta(\tau)^\top
            \right|_{\theta=\theta^\star}
        \right].
    \end{equation}
\end{proposition}

Any acceleration must therefore arise beyond this shared linearization. To study
that regime, we consider a finite set of complete trajectories with positive rewards,
a fixed backward policy, free logits for the forward trajectory probabilities $p_i$,
and a jointly learned normalizer $Z$. The target is
$p_i^\star=R(x_i)P_B(\tau_i\mid x_i)/Z^\star$, and trajectories are collected from
the trajectory-level mixture $Q_\varepsilon(\tau_i)=(1-\varepsilon)p_i+\varepsilon\mu_i$,
where $\varepsilon\in[0,1]$ controls exploration and $\mu$ is a fixed distribution
with full support. Even under this off-policy mixture, sign-selective feedback
preserves the target equilibrium.

\begin{theorem}[Fixed-point equivalence]
    \label{thm:fixed-point-equivalence}
    For every $\varepsilon\in[0,1]$, NRD and semi-gradient Trajectory Balance
    share a unique interior equilibrium, given by the target distribution
    $p^\star$ and normalizer $Z^\star$.
    \looseness=-1
\end{theorem}

Thus, NRD avoids the additional interior equilibria that can arise with the full
expected-loss gradient or Naive NRD. To examine progress toward the target, we measure
the spread of the trajectory log-ratios: writing $r_i=\log(p_i/p_i^\star)$, define
$D(p,p^\star)=\max_i r_i-\min_i r_i$. The normalizer adds the same offset to all
residuals, so the trajectories attaining these extrema also attain the largest
and smallest residuals. The NRD correction cannot increase their spread,
yielding the following pointwise comparison.
\looseness=-1

\begin{theorem}[Pointwise discrepancy comparison]
    \label{thm:pointwise-dominance}
    For every $\varepsilon\in[0,1]$, the discrepancy obeys
    $\dot D_{\mathrm{NRD}}\leq\dot D_{\mathrm{TB}}$ when both vector fields
    are evaluated at the same interior state.
    \looseness=-1
\end{theorem}

At a given distribution and normalizer, NRD therefore reduces the discrepancy
at least as fast as TB whenever TB is contracting it. This compares the methods
at the same state, rather than after they have followed different training paths,
and does not by itself guarantee contraction under arbitrary off-policy sampling.
On-policy, TB contracts $D$ whenever $D>0$, so NRD inherits this property.
The contraction keeps trajectory probabilities away from the boundary and yields
convergence of both the distribution and the normalizer. For the trajectory-level
mixture, the comparison is additionally strict when $0\leq\varepsilon<1$ and
the normalizer stays at its conditional stationary value,
$\mathbb{E}_{Q_\varepsilon}[\delta]=0$.
\looseness=-1

\begin{corollary}[On-policy convergence]
    \label{cor:convergence-strict}
    For $\varepsilon=0$, NRD converges from every interior initialization
    to $p(t)\to p^\star$ and $Z(t)\to Z^\star$. At every common interior
    state away from balance,
    $\dot D_{\mathrm{NRD}}\leq\dot D_{\mathrm{TB}}<0$.
\end{corollary}

\begingroup
\interlinepenalty=10000
To compare the time needed to approach the target, we next consider a regular
one-dimensional exponential family,
$p_\eta(\tau)=h(\tau)\exp\{\eta T(\tau)-A(\eta)\}$, with nonconstant sufficient
statistic $T$ and target parameter $\eta^\star$. Under on-policy sampling and
fast normalization, the NRD correction reinforces the direction of the TB update.
Here we measure progress by the Jeffreys divergence
$J(p,p^\star)=D_{\mathrm{KL}}(p\Vert p^\star)+D_{\mathrm{KL}}(p^\star\Vert p)$,
which remains finite between interior parameters even when $T$ is unbounded.
\looseness=-1
\par
\endgroup

\begin{theorem}[Hitting-time advantage]
    \label{thm:expfam-hitting}
    Start NRD and semi-gradient Trajectory Balance from the same
    $\eta_0\neq\eta^\star$, and let $T_{\mathrm{NRD}}(\epsilon)$ and
    $T_{\mathrm{TB}}(\epsilon)$ be the first times at which
    $J(p_\eta,p_{\eta^\star})\leq\epsilon$. Then
    $T_{\mathrm{NRD}}(\epsilon)<T_{\mathrm{TB}}(\epsilon)$ for every
    $0<\epsilon<J(p_{\eta_0},p_{\eta^\star})$.
\end{theorem}

\begin{figure}[t]
    \centering
    \includegraphics[width=\textwidth]{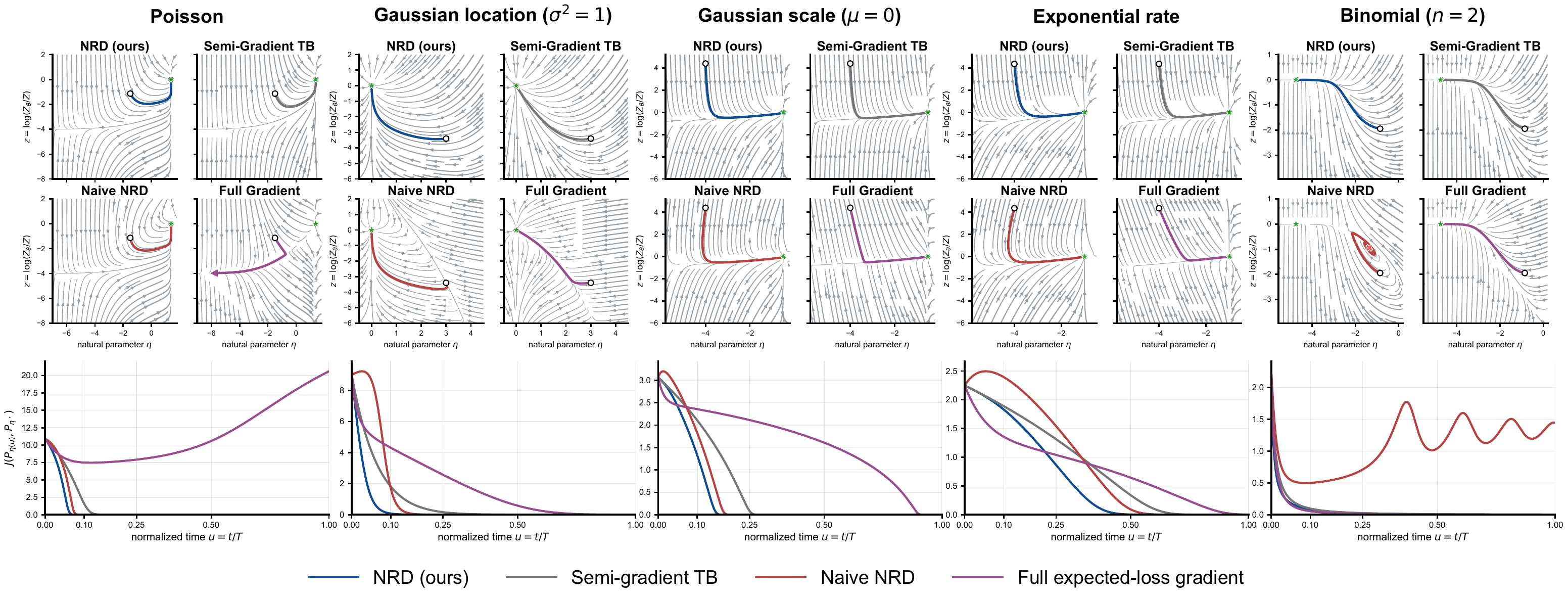}
    \vspace{-12pt}
    \caption{\textbf{Population dynamics in exponential families.}
        Top: phase portraits for Poisson, Gaussian location, Gaussian scale,
        exponential-rate, and binomial models. Bottom: Jeffreys divergence
        from the target along the illustrated trajectories. Time is normalized
        as $u=t/T\in[0,1]$ and displayed on a $\log(1+u)$ axis. These examples
        illustrate faster progress for NRD than semi-gradient TB and possible
    failure of Naive NRD or the full expected-loss gradient.}
    \label{fig:exponential-fam-dynamics}
\end{figure}

The two-trajectory model is the Bernoulli specialization of this result.
Its dynamics admit a closed form, and the additive hitting-time advantage
can grow without bound as the initial divergence increases.
Together, these results distinguish unchanged first-order behavior near balance
from stronger progress away from equilibrium, with a strict hitting-time advantage
under the one-dimensional assumptions. Figure~\ref{fig:exponential-fam-dynamics}
illustrates population dynamics in representative exponential families;
proofs and the two-trajectory specialization appear in
Appendix~\ref{app:theory-proofs}, with the latter in
Appendix~\ref{app:two-trajectory}.

\section{Experiments}
\label{sec:experiments}
\begingroup
\clubpenalties 3 10000 10000 0
\widowpenalties 3 10000 10000 0
\setcounter{topnumber}{2}
\setcounter{totalnumber}{3}

Our experiments test whether the behavior predicted by the theory holds in practice.
We ask whether \ours{} improves
distribution learning and high-reward discovery relative to TB, and whether
these gains require a separately trained exploratory policy. Comparing with
alternative training losses isolates the effect of the update rule; comparing
with auxiliary exploration methods tests how much exploration benefit survives
when sampler and GFlowNet are tied. We evaluate both target-distribution fit and
the discovery of distinct high-reward modes. \looseness=-1

\subsection{Experimental setup}
\label{sec:experimental-setup}

For distribution learning, we use Rings, Moons, and 8 Gaussians in DiffusionGrid,
adapted from the finite-horizon random-walk setting of
\citet{dallantonia2026boostedgflownetsimprovingexploration}, and the HyperGrid
construction benchmark
\citep{bengio2021flow,malkin2022trajectory}. Many paths lead to each terminal
state, but finite-budget exploration can still concentrate on only part of the
multimodal target. We measure estimated terminal total variation (TV) from
normalized reward. For high-reward discovery, we use Bags, Knapsack128, and
autoregressive sequence tasks adapted from prior structured benchmarks
\citep{silva2025when}, tracking the mean log-reward of the
top 200 unique objects encountered. We also study peptide design
\citep{jain2022biological,dallantonia2026avoid}, counting distinct sequences
that satisfy a screening criterion from an oracle trained on GRAMPA
\citep{witten2019deep}.

The first baseline group changes the training objective without adding an
exploration policy. It comprises TB, the asymmetric Linex(1) loss
\citep{hu2026squarederrorexploringloss}, and the pairwise Contrastive Balance
(CB) objective \citep{dasilva2024embarrassinglyparallelgflownets}. The second
group explicitly trains exploratory models through ACE's adversarial sampler
\citep{dallantonia2026avoid}, SA-GFN's novelty-seeking sibling policy
\citep{madan2025sibling}, or Adaptive Teacher's discrepancy-driven teacher
\citep{kim2025adaptive}.

We use four seeds and the prescribed budgets. Each such step counts a policy update, including
auxiliary-policy updates. Appendix~\ref{app:reproducibility} gives environment
definitions, architectures, optimization settings, estimators, and the complete
structured results. Appendix~\ref{app:experimental-context} provides baseline
context and wall-time curves; Appendix~\ref{app:score-ablation} contains the
score-distribution ablation.

\subsection{Distribution learning and exploration}
\label{sec:experiments-objectives}
\label{sec:experiments-auxiliary}

\begin{figure}[t]
    \centering
    \includegraphics[width=\textwidth]{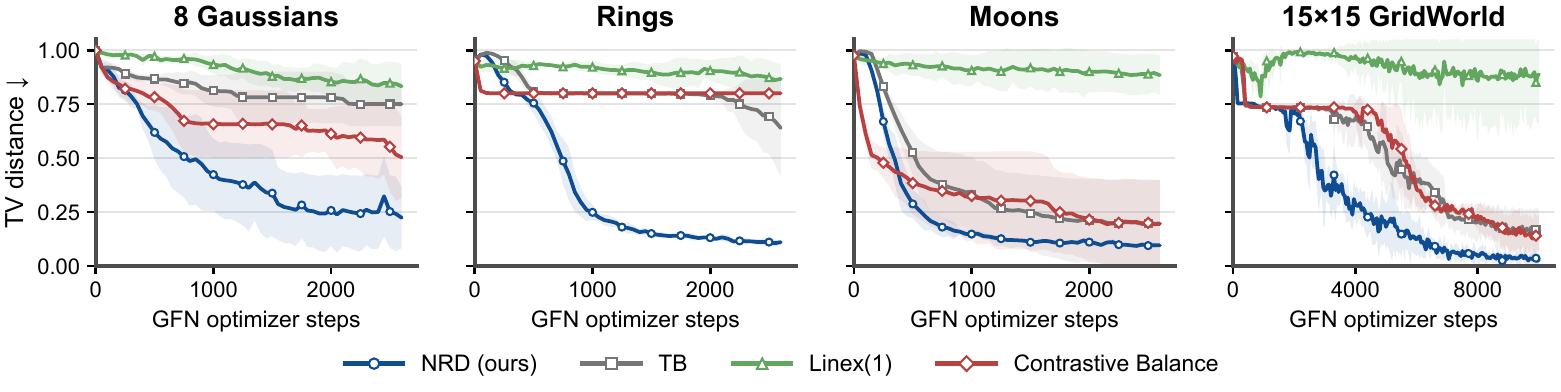}
    \vspace{-12pt}
    \caption{\textbf{Distribution learning with alternative training objectives.}
        Estimated terminal TV versus GFlowNet optimizer steps for \ours{}, TB,
        Linex(1), and Contrastive Balance. Lines and shaded bands show the mean
        and one standard deviation over four seeds. The GridWorld panel corresponds
    to the $15\times15$ HyperGrid. Lower values are better.}
    \label{fig:exp-objectives}
\end{figure}

\begin{figure}[!htb]
    \centering
    \includegraphics[width=\textwidth]{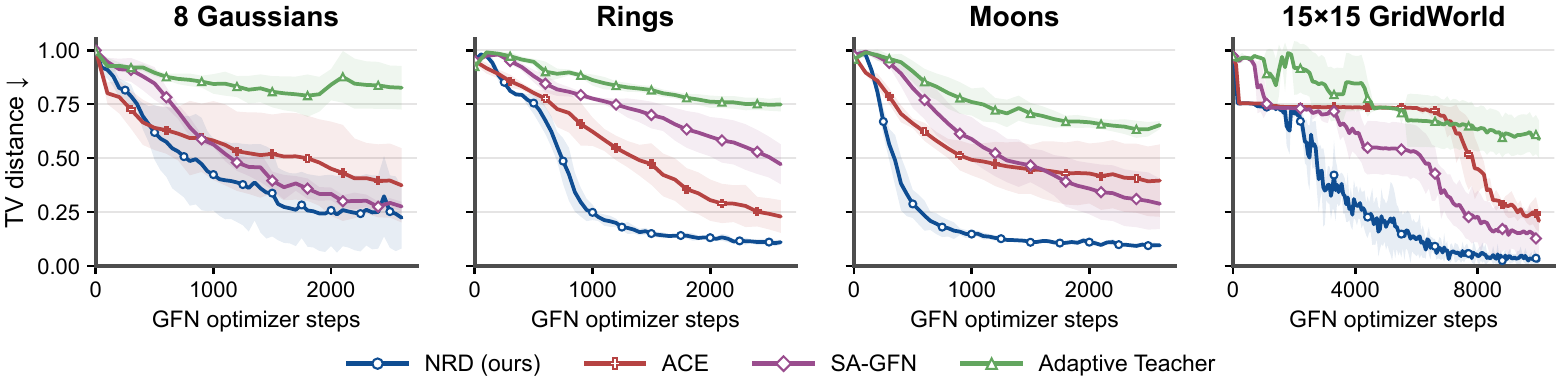}
    \vspace{-12pt}
    \caption{\textbf{Comparison with explicit exploration mechanisms.}
        \ours{} versus ACE, SA-GFN, and Adaptive Teacher on the same four
        distribution-learning tasks. Curves use the full optimizer-step budgets
        and report means with one-standard-deviation bands across four seeds.
    \ours{} has the lowest mean final estimated TV in each panel.}
    \label{fig:exp-auxiliary}
\end{figure}

Figures~\ref{fig:exp-objectives} and~\ref{fig:exp-auxiliary} address the same
exploration problem from two directions. First, can a single-policy update escape
the poor distribution fits observed under other GFlowNet losses? Rings provides
the clearest example. TB and CB remain at high error for substantial portions
of training, whereas \ours{} keeps improving, ending at estimated TV $0.110$
against TB's $0.639$. This illustrates the gap between a loss admitting the
correct distribution and its optimization reaching that distribution within a
finite budget. Across the four tasks, \ours{} has the lowest mean estimated TV. \looseness=-1

Second, is an independently trained exploration policy necessary for this
improvement? Figure~\ref{fig:exp-auxiliary} shows that auxiliary methods can
substantially reduce the distributional error seen in the first comparison,
but \ours{} remains competitive without a second sampler and has the lowest
estimated TV in each panel. SA-GFN is particularly competitive on
8 Gaussians, with overlapping seed variability. Together, these comparisons
support the proposed connection to adaptive exploration, with useful exploration
pressure surviving when the GFlowNet is responsible for correcting balance and sampling
trajectories rather than requiring an
additional policy. \looseness=-1

The optimizer-step comparison includes auxiliary-policy updates, while the
wall-time views in Appendix~\ref{app:experimental-context} expose the recorded
cost of training and online evaluation. The advantage on Rings and Moons also
appears over substantial shared portions of those time axes. The structured results below further
qualify the conclusion by showing cases where auxiliary exploration finds
higher-reward objects despite \ours{}'s stronger distribution-learning results.

\Needspace{36\baselineskip}
\subsection{Exploration with sparse rewards}
\label{sec:experiments-structured}
\label{sec:experiments-peptides}

\begin{wrapfigure}{r}{0.41\textwidth}
    \vspace{-1em}
    \centering
    \includegraphics[width=\linewidth]{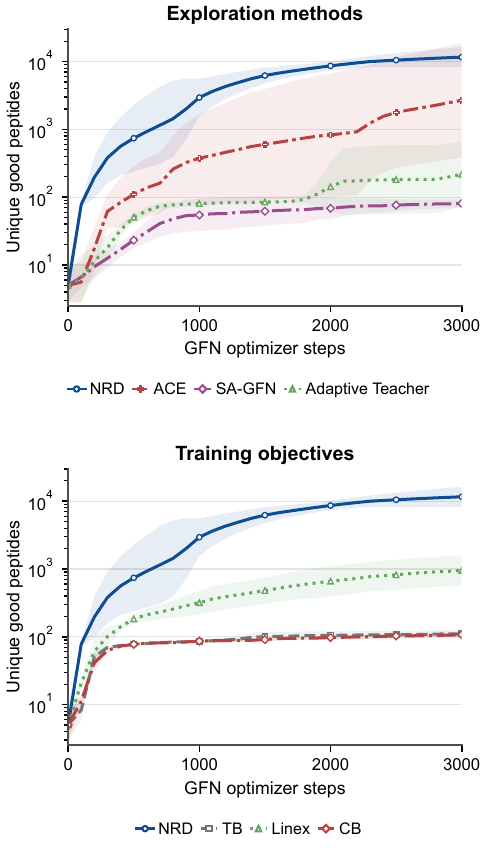}
    \vspace{-1.85em}
    \caption{\textbf{Cumulative peptide discovery.}
        Auxiliary exploration methods above; alternative objectives below.
    Geometric means over four seeds.}
    \vspace{-1.9em}
    \label{fig:exp-peptides}
\end{wrapfigure}

We next ask whether the exploration benefit translates into discovering distinct
peptide candidates among sequences of up to ten amino acids. The 19-amino-acid
alphabet already allows $19^{10}\approx6.1\times10^{12}$ full-length candidates.
Figure~\ref{fig:exp-peptides} accumulates oracle-qualified sequences sampled from
the principal policy without exploration noise; repeated discoveries do not
increase the count. A sequence qualifies when the maximum predicted activity
probability across five random forests is at least $0.95$. The evaluation
schedule, oracle, and aggregation protocol
are specified in Appendix~\ref{app:repro-peptides}. This measures the breadth
of computationally screened discoveries, not global distribution matching or
experimentally validated biological activity.
\looseness=-1

At the end of the budget, \ours{} discovers a geometric mean of $11{,}637$
distinct qualifying peptides, compared with $2{,}674$ for ACE, $953$ for Linex,
$218$ for Adaptive Teacher, $111$ for TB, $107$ for CB, and $81$ for SA-GFN.
The aggregate advantage is approximately $4.35\times$ over ACE and $105\times$
over TB. The trajectory of the count indicates that this advantage develops
during training, while the per-seed results show a narrower multiplicative
spread than ACE. The advantage is not seed-wise dominance, since ACE exceeds \ours{}
in one of the four seeds.

For the remaining structured tasks, we assess the quality of the discoveries
accumulated during training through the mean log-reward of the top 200 unique
objects. The complete results in Table~\ref{tab:structured-full} of
Appendix~\ref{app:reproducibility} show that \ours{} attains higher final means
than TB on Knapsack and all sequence tasks, while the Bags results are comparable within the
observed seed variability. \ours{} also remains competitive with auxiliary
exploration methods, including those driven by auxiliary rewards, although
some sequence tasks favor ACE or Adaptive Teacher. These results extend the
peptide findings to other combinatorial spaces, supporting effective
high-reward discovery without a separately trained exploration policy.
\par
\ifnum\value{WF@wrappedlines}>1
\vspace{\dimexpr\value{WF@wrappedlines}\baselineskip-2\baselineskip\relax}
\fi
\WFclear
\endgroup

\section{Related Work}
\label{sec:related-work}

Approaches to improving exploration in GFlowNet training follow two main
directions. The first changes the training objective while keeping a single
policy; the second learns an additional mechanism that redirects sampling toward
regions the GFlowNet has not yet learned.

Within the first direction, Linex(1) \citep{hu2026squarederrorexploringloss}
replaces the squared residual with a zero-avoiding loss designed to
encourage exploration and sample diversity, and $f$-Trajectory Balance
\citep{fawkes2026ftrajectorybalance} generalizes this loss design to the family
of $f$-divergences. Contrastive Balance
\citep{dasilva2024embarrassinglyparallelgflownets} compares the residuals of
pairs of trajectories so that the partition function cancels. With a closely
related objective, Stable-GFlowNet \citep{kwon2026stablegflownet}
links mode collapse to the variance of the estimated normalizer and shows
that removing it yields more diverse samples.

In the second direction, ACE \citep{dallantonia2026avoid} sets up a cooperative
two-player game in which an exploration GFlowNet samples proportionally to the
reward in regions under-allocated by the canonical GFlowNet. SA-GFN
\citep{madan2025sibling} trains a sibling GFlowNet whose reward combines the task
reward with a novelty bonus, and Adaptive Teacher \citep{kim2025adaptive} trains
a teacher GFlowNet rewarded by the student's loss and the task reward, steering
the student toward high-error regions. Boosted GFlowNets
\citep{dallantonia2026boostedgflownetsimprovingexploration} train a sequence of
GFlowNets, each fitting the reward not yet captured by its predecessors. These
methods make exploration adaptive, but at the cost of training and maintaining
models beyond the principal policy, along with additional hyperparameters such
as allocation thresholds and reward exponents.

\ours{} combines features of both directions. Like the methods in the first, it
trains a single policy without auxiliary models and introduces no hyperparameters
beyond those of Trajectory Balance; like those in the second, it
includes an explicit mechanism that directs training toward unresolved balance
violations. This mechanism, however, is obtained from the policy's own score
term rather than from a separately learned sampler or teacher, and it does not
arise from a choice of loss. Although we analyze and evaluate \ours{} with
Trajectory Balance, the same construction could in principle be applied on top
of other balance losses.

\section{Conclusion}
\label{sec:conclusion}

We introduced Negative Reinforce Dynamics, a novel GFlowNet training scheme that arises as a `weight-tied' version of an adversarial game between the GFlowNet and a trajectory sampler.
By selectively reinforcing under-allocated trajectories, NRD incorporates
the adversarial sampler's exploratory feedback into the model's own update, greatly enhancing discovery
while training a single model with no additional parameters. Our
experiments support the view that these exploratory benefits can survive parameter
tying, showing improved distribution learning and high-reward discovery relative to
Trajectory Balance, alongside competitive performance against auxiliary exploration
methods. NRD thus provides a way to exploit the adversarial formulation without
maintaining a separately trained sampler.

\subsection*{Acknowledgments}

We thank Emmanuel Bengio for valuable feedback on earlier drafts of this paper,
which improved its clarity and positioning. We acknowledge the support of the Funda\c{c}\~{a}o Carlos Chagas Filho de Amparo
\`{a} Pesquisa do Estado do Rio de Janeiro (FAPERJ) (SEI-260003/020348/2025,
SEI-260003/020694/2025) and the Conselho Nacional de Desenvolvimento
Cient{\'\i}fico e Tecnol\'{o}gico (CNPq) (404336/2023-0, 305692/2025-9,
445170/2024-7).

\bibliographystyle{iclr2027_conference}
\bibliography{references}

\appendix
\section{Reproducibility and complete structured-discovery results}
\label{app:reproducibility}

\subsection{Evaluation and optimization budgets}
We use seeds 39--42 and retain the prescribed final budget for each experiment,
without selecting a best checkpoint. Distribution-learning curves and the
structured-discovery tables report arithmetic means and sample standard
deviations across these four seeds. Peptide counts instead use geometric means
and one population standard deviation in log-count space. These bands describe
seed variability, not confidence intervals. An optimizer step in the grid and
peptide comparisons counts a GFlowNet-policy update: TB, NRD, Linex, and CB
contribute one per training iteration, whereas ACE, SA-GFN, and Adaptive Teacher
contribute two, including their auxiliary policy. The separate predictor update
used for SA-GFN's intrinsic reward is not a GFlowNet-policy update. Accordingly,
the DiffusionGrid budget is 2,600 iterations for the single-policy methods and
1,300 for auxiliary-policy methods; HyperGrid uses 10,000 and 5,000, respectively,
and peptides use 3,000 and 1,500. The structured tasks retain their original
iteration budgets, listed in Table~\ref{tab:structured-config}, for every method.

The grid metric is estimated terminal total variation,
$\widehat{\mathrm{TV}}=\tfrac12\sum_x|\hat p_\theta(x)-p^\star(x)|$,
where $p^\star(x)=R(x)/\sum_yR(y)$. Both terminal grids are enumerated, but
the forward marginal is reconstructed from ten backward trajectories per
terminal state. Specifically, samples from $P_B(\cdot\mid x)$ estimate
$P_T(x)=\mathbb E_{P_B(\cdot\mid x)}[P_F(\tau)/P_B(\tau\mid x)]$;
the resulting nonnegative weights are normalized over the terminal grid.
Thus these curves are Monte Carlo estimates, not exact forward-policy
marginals. Evaluations are recorded every 50 training iterations; the axes
convert iterations to the policy-update count above. For structured discovery,
a history queue retains the 200 highest-reward unique terminal objects seen
during training. The reported statistic is the arithmetic mean of their log
rewards, not the reward of a fresh evaluation batch.

\subsection{Distribution-learning environments and models}
\label{app:repro-grids}
DiffusionGrid adapts the finite-horizon lazy random walk of
\citet{dallantonia2026avoid}, with Rings and 8 Gaussians targets and an
additional Moons target. It starts at the origin of $\{-15,\ldots,15\}^2$ and takes
30 actions. Each action moves one unit in either coordinate direction or
stays in place; boundary-crossing actions are masked. Time is part of the
state, so spatial revisits still define an acyclic construction process.
Before boundary masking, the horizon admits $5^{30}$ action sequences for
only $31^2$ terminal locations, making path multiplicity substantial.
The Rings density is the sum of two radial Gaussian kernels with radii 3
and 12, radial standard deviation 1, and a common multiplier of 10.
The 8 Gaussians density sums eight isotropic Gaussian kernels of standard
deviation 1, centered at equally spaced points on a circle of radius 12.
Moons sums kernels of standard deviation 1 over 256 equally spaced anchors
on each of two semicircles: their radius is 9, horizontal shifts are $-4.5$
and $4.5$, and the lower semicircle is shifted down by 2.7. For each density
$d(x)$, the reward is $R(x)=e[(1-10^{-6})d(x)+10^{-6}]$, up to a negligible
numerical floor inside the kernel sum. The common reward exponent is one.

The principal DiffusionGrid forward policy has two hidden layers of width
128 with ReLU activations and five output logits. The separately learned
backward policy is a linear map to five logits. Both receive spatial
coordinates, normalized time, and sine/cosine time features at eight
frequencies $2^k$, $k=0,\ldots,7$. There is no shared forward--backward
backbone. The ACE auxiliary networks use one hidden layer of width 64,
with 16 forward and eight backward time frequencies. Other auxiliary
policies follow the principal-policy dimensions. Training uses a constant
action-wise exploration probability $\varepsilon=0.1$ and base batch size
1,024. TB, NRD, CB, and the auxiliary-policy baselines use AdamW with policy
learning rates $0.005$, normalizer learning rate $0.05$ where applicable,
and weight decay $0.01$. The plotted Linex configuration uses AdamW with
forward/backward learning rates $0.001$ and normalizer learning rate $0.1$.
These rates are constant, without a learning-rate schedule or gradient clipping.

HyperGrid follows the coordinate-increment construction of
\citet{bengio2021flow,malkin2022trajectory}, with states
$x\in\{0,\ldots,14\}^2$. Starting at $(0,0)$,
the policy increments one coordinate or terminates, with invalid increments
masked. Let $a_i=|2x_i/14-1|$. Its reward is
$R(x)=10^{-3}+3\prod_{i=1}^2\mathbf 1\{0.6<a_i<0.8\}$.
The policy receives the coordinate vector and uses three hidden layers
of width 128 with leaky-ReLU activations, followed by three action logits.
Its backward policy is fixed and uniform over valid parents. Batch size
is 512 and $\varepsilon=0.1$. Matrix-valued policy parameters use Muon
and non-matrix parameters use its Adam-based branch, without weight decay
or gradient clipping. The network learning rate increases linearly from
zero to $0.01$ over the first 10\% of iterations, then follows a cosine
decay to $0.0001$ at the budget endpoint. The Linex curve uses the same
optimizer and schedule with peak rate $0.001$ and final rate $0.00001$.
The learned scalar $\log Z$ is initialized at zero and updated with Adam
at constant rate $0.1$. DiffusionGrid also initializes $\log Z$ at zero;
CB eliminates the normalizer from its paired-residual objective.

\subsection{Structured environments, optimization, and full results}
\label{app:repro-structured}
The structured tasks adapt the bag, quadratic-knapsack, and autoregressive
sequence constructions used by \citet{dallantonia2026avoid}; the sequence
family also follows \citet{silva2025when}. The definitions below specify our
instances and rewards rather than assuming identical published configurations.
Bags18 and Bags24 construct multisets of exactly 12 items from 18 and
24 item types, respectively; repeated items are permitted. A state is
its count vector, and each action adds one item. With positive lognormal
utilities $u_i$ generated using reward seed 43, the log reward is
$\log R(x)=10[\log(u^\top x)-c]$, where
$c=\log\lfloor 12\max_i u_i/2\rfloor$ is the common normalizing offset.
The backward policy removes one item type uniformly among types with
positive count, rather than sampling individual copies proportionally
to their counts.

Knapsack128 constructs a subset of 128 items, with each item selectable
at most once and total weight at most 60. Integer weights are sampled
uniformly from $\{2,\ldots,58\}$ for each run seed and shared by methods
at that seed. Construction stops when no remaining item fits. The reward
is $R(x)=u^\top x+\tfrac12x^\top A x$. Using reward seed 43, the utilities
are uniform on $[0.1,1)$, and the symmetric interaction matrix is formed
from standard-normal off-diagonal entries with sparsity 0.15. Each diagonal
entry is the absolute off-diagonal row sum plus 0.1; entries are then
divided by $\sqrt{w_iw_j}+10^{-6}$. The backward policy removes a selected
item uniformly. The weight realization varies across seeds, while methods
within a seed face the same instance.

The sequence environments append tokens autoregressively to a fixed
length $L$, so the backward transition is deterministic. Binary32 uses
$L=32$ and vocabulary size $V=2$; the remaining tasks use each combination
of $L\in\{24,32\}$ and $V\in\{4,6\}$. All five tasks use the additive
log reward $\log R(x)=\sum_{i=1}^{L}(u_{x_i}+v_i)$, with token utilities
$u$ and position offsets $v$ sampled uniformly from $[0,1)$ using reward
seed 43. In particular, Binary32 here uses this reward, not a separate
Hamming-distance-to-modes objective. Network inputs are zero-padded prefix
vectors for sequences and item-count vectors for Bags and Knapsack.

All structured-task forward policies use three hidden layers of width
128 with leaky-ReLU activations (negative slope 0.01), followed by action
logits. Backward policies are fixed as described above. The policy optimizer,
10\%-warmup cosine schedule, scalar normalizer initialization, and separate
normalizer learning rate are those of the standard HyperGrid configuration.
Thus the policy peak learning rate is $0.01$ and the constant Adam rate
for $\log Z$ is $0.1$; no weight decay or gradient clipping is applied.
Auxiliary policies use the same architecture and base learning rates.
The exploration probability is constant at 0.1, mixing with a uniform
distribution over valid actions at each step. ACE and SA-GFN split the
configured base batch equally between their two rollout branches, while
TB, NRD, and Adaptive Teacher use the full base batch per rollout.

\begin{table}[!htb]
    \centering
    \caption{Structured-task configurations. Each budget is a number of
    training iterations for every method; checkpoint intervals use the same unit.}
    \label{tab:structured-config}
    \small
    \setlength{\tabcolsep}{9pt}
    \begin{tabular}{lrrrr}
        \toprule
        Environment & Size / vocabulary & Batch & Iterations & Checkpoint interval\\
        \midrule
        Bags18 & $18/12$ & 32 & 1,500 & 50\\
        Bags24 & $24/12$ & 32 & 1,500 & 50\\
        Knapsack128 & $128/60$ & 32 & 256 & 10\\
        Binary32 & $32/2$ & 64 & 2,000 & 50\\
        Seq24/V4 & $24/4$ & 64 & 2,000 & 50\\
        Seq24/V6 & $24/6$ & 64 & 2,000 & 50\\
        Seq32/V4 & $32/4$ & 64 & 2,000 & 50\\
        Seq32/V6 & $32/6$ & 64 & 2,000 & 50\\
        \bottomrule
    \end{tabular}
\end{table}

\begin{table}[!htb]
    \centering
    \caption{Complete structured-discovery results at the budgets in
    Table~\ref{tab:structured-config}. Entries are the mean log reward of
    the top 200 unique objects, averaged over four seeds, $\pm$ one sample
    standard deviation. Higher is better; bold marks means whose $\pm1$ SD
    intervals overlap that of the highest mean, using unrounded values.
    This describes seed variability, not statistical equivalence.}
    \label{tab:structured-full}
    \small
    \setlength{\tabcolsep}{3.7pt}
    \renewcommand{\arraystretch}{1.12}
    \begin{tabular}{lccccc}
        \toprule
        Environment & NRD & TB & ACE & SA-GFN & Adaptive Teacher\\
        \midrule
        Bags18 & $\mathbf{5.734}\pm0.008$ & $\mathbf{5.748}\pm0.038$ & $5.641\pm0.049$ & $\mathbf{5.726}\pm0.023$ & $\mathbf{5.719}\pm0.070$\\
        Bags24 & $\mathbf{5.459}\pm0.020$ & $\mathbf{5.468}\pm0.044$ & $5.260\pm0.075$ & $\mathbf{5.395}\pm0.073$ & $\mathbf{5.455}\pm0.077$\\
        Knapsack128 & $\mathbf{3.375}\pm0.093$ & $\mathbf{3.219}\pm0.098$ & $\mathbf{3.235}\pm0.070$ & $3.137\pm0.064$ & $3.191\pm0.081$\\
        Binary32 & $38.897\pm0.073$ & $38.483\pm0.028$ & $\mathbf{39.010}\pm0.497$ & $38.488\pm0.016$ & $\mathbf{39.306}\pm0.292$\\
        Seq24/V4 & $\mathbf{31.667}\pm0.107$ & $31.248\pm0.040$ & $\mathbf{32.105}\pm0.580$ & $31.221\pm0.037$ & $\mathbf{31.582}\pm0.045$\\
        Seq24/V6 & $\mathbf{30.244}\pm0.199$ & $29.781\pm0.073$ & $\mathbf{30.252}\pm0.265$ & $29.781\pm0.049$ & $\mathbf{29.968}\pm0.020$\\
        Seq32/V4 & $\mathbf{40.751}\pm0.286$ & $39.640\pm0.048$ & $\mathbf{42.011}\pm1.117$ & $39.626\pm0.083$ & $39.865\pm0.086$\\
        Seq32/V6 & $\mathbf{38.923}\pm0.577$ & $37.504\pm0.139$ & $\mathbf{39.997}\pm0.913$ & $37.537\pm0.104$ & $37.539\pm0.130$\\
        \bottomrule
    \end{tabular}
\end{table}

Table~\ref{tab:structured-full} reports all structured tasks discussed in
Section~\ref{sec:experiments-structured}. TB has the highest mean on both Bags tasks, NRD on Knapsack,
Adaptive Teacher on Binary32, and ACE on all four nonbinary sequence tasks.
The small mean differences in Bags and Seq24/V6 should be read together
with their seed variability; these endpoint rankings are descriptive and
are not significance tests.

\subsection{Peptide oracle, model, and discovery protocol}
\label{app:repro-peptides}
Our peptide task follows the oracle-guided sequence-design setting
\citep{jain2022biological,dallantonia2026avoid}, with the alphabet and
screening protocol specified here. Peptides are generated by appending
amino acids or an end-of-sequence token,
with maximum length ten. The alphabet contains the 19 standard amino acids
other than cysteine; the end-of-sequence token also provides padding. The
policy embeds each token in 32 dimensions, concatenates an eight-token context
window with four sinusoidal position features, and applies one 64-unit ReLU
hidden layer before the output logits. The backward transition is deterministic.
The base batch is 4,096 and the constant exploration probability is 0.3.
NRD, TB, CB, ACE, SA-GFN, and Adaptive Teacher use AdamW with network rate
$0.05$ and weight decay $0.01$; learned log normalizers use rate $0.1$.
Linex uses Adam, no weight decay, network rate $0.005$, and normalizer rate
$0.02$. There is no learning-rate schedule or gradient clipping. ACE and
SA-GFN collect 2,048 trajectories per branch. All methods use the same
seed-specific initial principal-policy checkpoint.

The oracle consists of five random-forest classifiers trained on filtered
GRAMPA records \citep{witten2019deep}, one for each of \emph{E. coli},
\emph{S. aureus}, \emph{P. aeruginosa}, \emph{B. subtilis}, and
\emph{C. albicans}. The preprocessing retains sequences of length one to ten,
excludes cysteine and specified chemical modifications, and removes records
from YADAMP. Listed peptides associated with each organism provide positive
examples; an equal number of randomly generated sequences, using the empirical
positive-length distribution, provide negative examples. The respective
positive/negative counts per classifier are 664/664, 653/653, 396/396,
170/170, and 227/227. Classifiers use flattened one-hot sequence features,
500 trees, square-root feature subsampling, balanced class weights, and
random seed 42. These models provide computational screening scores, not
experimental validation of generated candidates.

Let $p(x)$ be the maximum positive-class probability across the five
classifiers, and let $\ell(x)$ be the sequence length. The training log reward is
\begin{equation}
    \log R(x)=\max\!\left\{-30,
      (\ell(x)+1)\min\!\left[0,
      \frac{\operatorname{logit}p(x)-\operatorname{logit}(0.95)}{0.3}
      \right]\right\}.
\end{equation}
This transformation is bounded above by zero, with reward one for qualifying
sequences; its clipping is part of the reward definition. A discovery requires
$p(x)\geq0.95$. Evaluation samples 1,000 sequences with zero exploration
noise at initialization and every 100 GFlowNet-policy updates through 3,000.
Sequences are deduplicated cumulatively within each run, including those from
the initial evaluation. The reported $11{,}637$ is the geometric mean of
these cumulative counts at step 3,000, not a count from a single batch or
a sum over seeds. Generation enforces the vocabulary and length bound;
no additional biological-validity filter is applied to the discovery count.

\section{Connections to exploration baselines and wall-time comparisons}
\label{app:experimental-context}

\subsection{GFlowNet losses without an auxiliary exploration policy}

Trajectory Balance (TB) \citep{malkin2022trajectory} trains a generative policy
by matching the forward trajectory flow to the terminal reward times the
backward trajectory probability. Its squared log-residual penalty provides the
reference update in our comparisons. Changing this penalty can change the
training dynamics without changing the desired balance relation. In particular,
Linex(1) \citep{hu2026squarederrorexploringloss} is asymmetric: with our residual
convention, its implemented penalty is $\exp(-\delta)-1+\delta$. Negative
residuals therefore receive different weights from equally large positive ones.
This is distinct from \ours{}, which retains the TB semi-gradient and adds a
detached, negative-residual score correction to the policy update.

Contrastive Balance \citep{dasilva2024embarrassinglyparallelgflownets} compares
trajectory log-ratios pairwise. The evaluated implementation minimizes squared
differences between paired TB residuals, so the shared log-normalizer cancels.
Like TB and Linex, it does not train a separate exploration policy. These
comparisons distinguish a change in the balance loss from the additional
sampling-distribution pressure in \ours{}; they do not exhaust all possible
losses or their hyperparameter settings. Correctness of a balance relation at
its solution is also separate from finite-budget coverage: a full-support
policy can assign practically negligible probability to important regions long
before it reaches that solution.

\subsection{Methods with an auxiliary exploration policy}

ACE \citep{dallantonia2026avoid} couples a canonical GFlowNet with an
exploration GFlowNet in what its authors describe as a cooperative two-player
game: the explorer samples proportionally to the reward in regions
under-allocated by the canonical GFlowNet, which is trained on trajectories
from both policies. Its two-player
structure is especially relevant here: \ours{} investigates whether an
exploration-inducing update can be retained after identifying sampler and
learner parameters. ACE instead maintains the additional sampler throughout
training. The comparison tests the practical effect of this distinction, not
an equivalence between their finite-step implementations.

SA-GFN \citep{madan2025sibling} augments the principal GFlowNet with a sibling
policy that uses an intrinsic novelty signal. In the evaluated implementation,
random-network-distillation prediction error supplies this signal, and the
principal policy learns from sibling-generated trajectories as well as its own.
This adds both a policy and a learned novelty predictor. Adaptive Teacher
\citep{kim2025adaptive} instead constructs a teacher reward from the student's
trajectory-balance discrepancy and the task reward, emphasizing trajectories
on which the student needs improvement. The teacher and student are optimized
separately. Both approaches make exploration adaptive, but they require learned
components beyond the principal policy, unlike \ours{}.

\subsection{Recorded wall-time comparisons}
\label{app:walltime}

\begin{figure}[t]
    \centering
    \includegraphics[width=\textwidth]{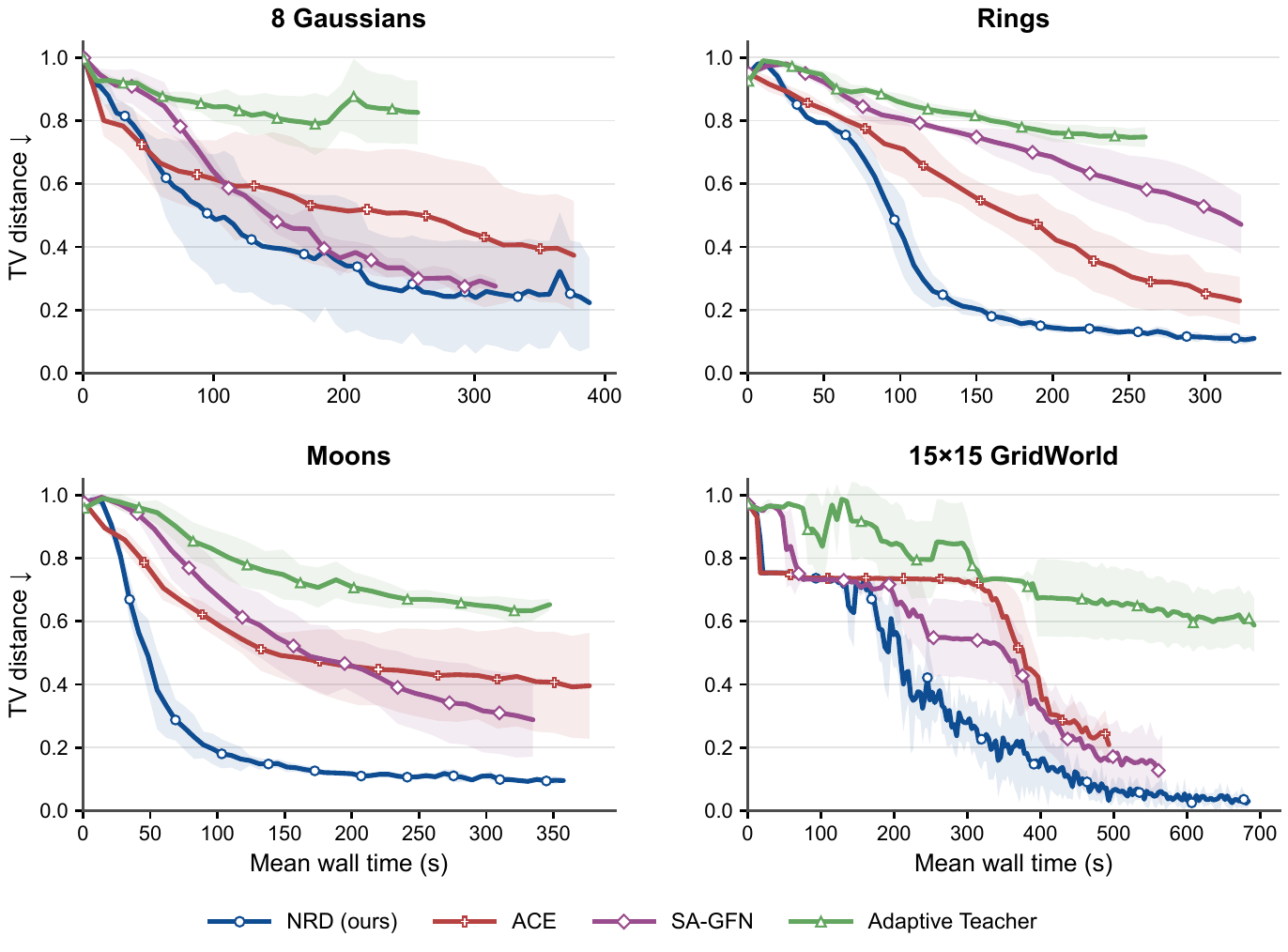}
    \caption{\textbf{Distribution learning versus recorded wall time.}
        Estimated TV for the auxiliary-policy comparison in
        Figure~\ref{fig:exp-auxiliary}, using the same source runs and seeds
        39--42. Each point pairs mean elapsed time with mean TV at the matching
        checkpoint; shading is one sample standard deviation in TV. Elapsed
        time includes training and online evaluation. Curves retain their
        recorded endpoints, which differ across methods. Lower is better.}
    \label{fig:exp-walltime}
\end{figure}

Figure~\ref{fig:exp-walltime} re-expresses the distribution-learning comparison
against elapsed time rather than GFlowNet-policy updates. On Rings and Moons,
the lower estimated TV of \ours{} remains visible over substantial shared
portions of the recorded time range. The 8 Gaussians comparison is less
separated from SA-GFN and has appreciable seed variability. HyperGrid also shows
earlier progress by \ours{} after the initial plateau, but methods finish at
different times. Thus the benefit is not visible only under the policy-step
accounting, while neither the plots nor their unequal endpoints establish a
universal wall-time advantage.

Elapsed time is the recorded cumulative wall time, or the sum of cumulative
training and online-evaluation time where a wall-time field is unavailable.
Aggregation is by checkpoint, not by interpolation onto a common time grid:
horizontal positions are the means of the four seed times. The figure therefore
describes the recorded error--time trade-off rather than an equal-time
stopping rule.

\section{Sensitivity to the exploration distribution}
\label{app:score-ablation}

We study how the score distribution affects training when trajectories are
collected with an action-wise $\varepsilon$-greedy mixture
$Q_\varepsilon(a\mid s)=(1-\varepsilon)P_F(a\mid s)+\varepsilon U(a\mid s)$,
where $U$ is uniform over valid actions. The ablation compares the corrections
based on $\nabla\log P_F$ and $\nabla\log Q_\varepsilon$, keeping the remaining
settings fixed within each environment and noise level. Four seeds are used
for each pair, with $\varepsilon\in\{0.1,0.3,0.5,0.8\}$. We summarize entire
learning curves rather than only the final checkpoint, since two variants may
reach similar final errors while differing substantially in their transients.
This ablation is separate from the main baseline comparisons.

For an error curve $m(t)$, we compute the normalized area over its prescribed
budget $T$ and report the relative reduction obtained with $Q_\varepsilon$:
\begin{equation}
    A_s=\frac{1}{T}\int_0^T \overline{m_s(t)}\,dt,
    \qquad
    \Delta_{\mathrm{AUC}}=100\left(1-\frac{A_Q}{A_{P_F}}\right).
    \label{eq:score-auc}
\end{equation}
The integral is evaluated by the trapezoidal rule at recorded checkpoints,
and the bar denotes the arithmetic mean across seeds. Positive values indicate
lower cumulative error with $Q_\varepsilon$; negative values indicate the
opposite. The grids use estimated TV. For the structured tasks, we use a
batch-normalized implementation of Flow Consistency in Subgraphs (FCS)
\citep{silva2025when}, averaged over evaluation batches:
terminal states are sampled by a uniform-action rollout, model and reward
weights are normalized within each batch, and their TV is computed. FCS is
therefore a consistency proxy, not the exact TV over the entire terminal space
and not the historical top-200 reward metric used in
Section~\ref{sec:experiments-structured}.

\begin{table}[!htb]
    \centering
    \small
    \setlength{\tabcolsep}{12pt}
    \caption{\textbf{Percentage reduction in error-curve area from using
        $Q_\varepsilon$ rather than $P_F$ in the score correction.}
        Positive values favor $Q_\varepsilon$. All entries aggregate four seeds
        over the full prescribed budget; they are effect sizes, not significance
    tests. The two blocks use different error metrics.}
    \label{tab:score-auc}
    \begin{tabular}{lrrrr}
        \toprule
        Environment & $\varepsilon=0.1$ & $0.3$ & $0.5$ & $0.8$ \\
        \midrule
        \multicolumn{5}{l}{\emph{Estimated terminal TV}}\\
        Rings       & $-11.14$ & $ 2.50$ & $ 3.88$ & $-22.96$ \\
        Moons       & $-34.92$ & $-2.14$ & $-6.13$ & $-10.25$ \\
        8 Gaussians & $-13.93$ & $14.03$ & $31.43$ & $  3.71$ \\
        HyperGrid   & $  6.30$ & $11.40$ & $-1.86$ & $ -0.20$ \\
        \midrule
        \multicolumn{5}{l}{\emph{FCS}}\\
        Bags18      & $ 1.44$ & $ 5.89$ & $14.12$ & $20.75$ \\
        Bags24      & $ 0.87$ & $ 8.67$ & $10.89$ & $15.24$ \\
        Knapsack128 & $-1.06$ & $-1.88$ & $ 2.85$ & $ 8.00$ \\
        Binary32    & $42.26$ & $45.50$ & $32.66$ & $ 7.60$ \\
        Seq24/V4    & $17.13$ & $25.96$ & $18.64$ & $ 6.91$ \\
        Seq24/V6    & $ 6.32$ & $18.87$ & $16.82$ & $ 5.52$ \\
        Seq32/V4    & $23.08$ & $33.09$ & $25.95$ & $ 8.60$ \\
        Seq32/V6    & $16.71$ & $25.06$ & $21.83$ & $ 6.41$ \\
        \bottomrule
    \end{tabular}
\end{table}

Table~\ref{tab:score-auc} shows that the effect depends on both the task and the
exploration level. In the structured tasks, the $Q_\varepsilon$ correction
reduces area in 30 of the 32 environment--noise combinations, with the largest
reduction reaching $45.50\%$ on Binary32 at $\varepsilon=0.3$. Knapsack is
the exception at the two lowest noise levels, where the differences favor
$P_F$ by less than two percent. The grid results are more mixed: only seven
of the sixteen combinations favor $Q_\varepsilon$, and Moons favors $P_F$
at every evaluated noise level. These observations establish sensitivity to
the score distribution and substantial transient benefits in many structured
tasks, but do not support a universal ordering of the two variants.

\section{Proofs of the theoretical results}
\label{app:theory-proofs}
This section first establishes the marginalization and gradient identities
used in the main text, and then proves the results in Section~\ref{sec:theory}.
We use the same notation throughout and introduce additional notation only
when needed.

\subsection{Trajectory balance implies the target terminal marginal}
\label{app:tb-marginal}
Assume complete forward trajectories have total probability one, and the
backward policy defines a normalized distribution over trajectories ending
at each terminal object. If Equation~\ref{eq:tb} holds for every complete
trajectory, summing over all trajectories ending at a fixed $x$ gives
\begin{equation}
    ZP_T(x)
    =Z\sum_{\tau\to x}P_F(\tau)
    =R(x)\sum_{\tau\to x}P_B(\tau\mid x)
    =R(x).
\end{equation}
Summing this identity over $x$ yields $Z=\sum_x R(x)=Z^\star$, because
$\sum_xP_T(x)=1$. Substitution gives $P_T(x)=R(x)/Z^\star$ for every
terminal object. This is the usual marginalization consequence of trajectory
balance \citep{malkin2022trajectory,bengio2023foundations}.

\subsection{Score-function decomposition of the expected-loss gradient}
\label{app:score-proof}
For Equations~\ref{eq:intro-score-decomposition} and
\ref{eq:score-decomposition}, assume a parameter-independent trajectory
support on which $Q_\theta$ is positive and differentiable, and that
derivatives can be interchanged with the expectation. On a finite support,
the product rule directly gives
\begin{equation}
    \begin{aligned}
        \nabla_\theta\sum_\tau Q_\theta(\tau)L_\theta(\tau)
        &=\sum_\tau Q_\theta(\tau)\nabla_\theta L_\theta(\tau)
        +\sum_\tau L_\theta(\tau)\nabla_\theta Q_\theta(\tau)\\
        &=\mathbb E_{Q_\theta}[\nabla_\theta L_\theta]
        +\mathbb E_{Q_\theta}
        [L_\theta\nabla_\theta\log Q_\theta],
    \end{aligned}
\end{equation}
where the second line uses
$\nabla_\theta Q_\theta=Q_\theta\nabla_\theta\log Q_\theta$.
The same argument applies to densities when differentiation under the
integral is justified. This is the classical score-function identity
\citep{williams1992simple}; the semi-gradient retains only its first term.

\subsection{The TB semi-gradient and reverse KL}
\label{app:tb-kl-proof}
We derive Equation~\ref{eq:tb-reverse-kl}, following the identity established
by \citet{whitammer2026gflownetsvariationalinference}. Let
$p_\theta(\tau)=P_F^\theta(\tau)$ and
$\pi(\tau)=R(x)P_B(\tau\mid x)/Z^\star$, with the backward policy, reward,
and $Z$ held fixed under differentiation with respect to $\theta_F$.
The normalization of the backward policy makes $\pi$ a probability
distribution. Assume common positive support and the regularity needed
to differentiate expectations. Write
$\delta_\theta=\log(p_\theta/\pi)+c$, where
$c=\log Z-\log Z^\star$ is independent of $\theta_F$.
The score $s_\theta=\nabla_{\theta_F}\log p_\theta$ has zero mean,
since $\mathbb E_{p_\theta}[s_\theta]
=\nabla_{\theta_F}\sum_\tau p_\theta(\tau)=0$. Therefore
\begin{equation}
    \begin{aligned}
        \mathbb E_{p_\theta}[\nabla_{\theta_F}\delta_\theta^2]
        &=2\mathbb E_{p_\theta}
        [\{\log(p_\theta/\pi)+c\}s_\theta]
        =2\mathbb E_{p_\theta}[\log(p_\theta/\pi)s_\theta],\\
        \nabla_{\theta_F}D_{\mathrm{KL}}(p_\theta\Vert\pi)
        &=\mathbb E_{p_\theta}
        [\{\log(p_\theta/\pi)+1\}s_\theta]
        =\mathbb E_{p_\theta}[\log(p_\theta/\pi)s_\theta].
    \end{aligned}
\end{equation}
Combining these equalities proves the factor of two in
Equation~\ref{eq:tb-reverse-kl}. The identity does not require
$Z=Z^\star$: its constant offset cancels by the zero-mean score property.
The backward policy may be learned through separate parameters, but shared
forward--backward parameters would introduce additional derivative terms.

\subsection{Proof of Proposition~\ref{prop:local-dynamics}}
Write the semi-gradient TB and NRD population dynamics as
\begin{equation}
    \begin{aligned}
        \dot\theta_{\mathrm{TB}}
        &=
        -2\,
        \mathbb{E}_{\tau\sim Q_\theta}
        \left[
            \delta_\theta(\tau)
            \nabla_\theta\delta_\theta(\tau)
        \right],
        \\
        \dot\theta_{\mathrm{NRD}}
        &=
        \dot\theta_{\mathrm{TB}}
        +
        \mathbb{E}_{\tau\sim Q_\theta}
        \left[
            L_\theta^-(\tau)
            \nabla_\theta\log Q_\theta(\tau)
        \right].
    \end{aligned}
    \label{eq:app-local-fields}
\end{equation}
At a balanced solution $\theta^\star$,
$\delta_{\theta^\star}(\tau)=0$ for every trajectory. Hence both integrands
in Equation~\ref{eq:app-local-fields} vanish at $\theta^\star$.
Differentiating the TB field at this point, every term arising from the
variation of $Q_\theta$ or from the Hessian of $\delta_\theta$ is multiplied
by the residual and therefore vanishes. We obtain
\begin{equation}
    \left.
    \frac{\partial\dot\theta_{\mathrm{TB}}}
    {\partial\theta}
    \right|_{\theta^\star}
    =
    -2\,
    \mathbb{E}_{\tau\sim Q_{\theta^\star}}
    \left[
        \left.
        \nabla_\theta\delta_\theta(\tau)
        \nabla_\theta\delta_\theta(\tau)^\top
        \right|_{\theta=\theta^\star}
    \right].
\end{equation}
For the NRD correction, the scalar function
$x\mapsto x^2\,\mathbf{1}\!\{x<0\}$ is continuously differentiable at the
origin, with both its value and first derivative equal to zero. Consequently,
the derivative of the complete score correction also vanishes at
$\theta^\star$. The two Jacobians are therefore equal, which proves the
result.
\looseness=-1
\subsection{Proof of Theorem~\ref{thm:fixed-point-equivalence}}
Let
\begin{equation}
    q_i
    =
    (1-\varepsilon)p_i+\varepsilon\mu_i,
    \qquad
    \bar\delta_q
    =
    \sum_i q_i\delta_i,
    \qquad
    \overline{L^-}_p
    =
    \sum_i p_iL_i^-.
\end{equation}
Since the forward distribution is parameterized by free logits $\Phi$,
\begin{equation}
    \frac{\partial\log p_i}{\partial\Phi_j}
    =
    \mathbf{1}\!\{i=j\}-p_j.
\end{equation}
The exact-expectation dynamics of TB and NRD are therefore
\begin{equation}
    \begin{aligned}
        \dot{\log Z}
        &=
        -2\bar\delta_q,
        \\
        \dot\Phi_i^{\mathrm{TB}}
        &=
        -2\left(
            q_i\delta_i-p_i\bar\delta_q
        \right),
        \\
        \dot\Phi_i^{\mathrm{NRD}}
        &=
        -2\left(
            q_i\delta_i-p_i\bar\delta_q
        \right)
        +(1-\varepsilon)p_i
        \left(
            L_i^--\overline{L^-}_p
        \right).
    \end{aligned}
    \label{eq:app-mixture-dynamics}
\end{equation}
The last term follows from
\begin{equation}
    \nabla_\Phi\log q_i
    =
    \frac{(1-\varepsilon)p_i}{q_i}
    \nabla_\Phi\log p_i.
\end{equation}
If Trajectory Balance holds on every trajectory, then
$\delta_i=L_i^-=0$ for all $i$, so all three equations in
Equation~\ref{eq:app-mixture-dynamics} vanish. Every balanced solution is
therefore a fixed point of both dynamics.
Conversely, consider a fixed point of NRD. The normalizer equation gives
$\bar\delta_q=0$. If the residuals were not all zero, the full support of
$q$ would imply the existence of an index $m$ with $\delta_m>0$. For this
trajectory $L_m^-=0$, and Equation~\ref{eq:app-mixture-dynamics} gives
\begin{equation}
    \dot\Phi_m^{\mathrm{NRD}}
    =
    -2q_m\delta_m
    -(1-\varepsilon)p_m\overline{L^-}_p
    <0,
\end{equation}
contradicting stationarity. Hence $\delta_i=0$ for every $i$.
The same conclusion follows immediately for TB, since at a fixed point
$\bar\delta_q=0$ and
\begin{equation}
    0
    =
    \dot\Phi_i^{\mathrm{TB}}
    =
    -2q_i\delta_i.
\end{equation}
Because $q_i>0$, every residual must vanish.
Finally, $\delta_i=0$ implies
\begin{equation}
    Zp_i
    =
    R(x_i)P_B(\tau_i\mid x_i).
\end{equation}
Summing over all trajectories and using the normalization of
$P_B(\cdot\mid x)$ gives
\begin{equation}
    Z
    =
    \sum_i
    R(x_i)P_B(\tau_i\mid x_i)
    =
    \sum_xR(x)
    =
    Z^\star.
\end{equation}
It follows that $p_i=p_i^\star$ for every $i$. Thus the two dynamics have
the same unique fixed point in $(p,Z)$ coordinates:
\begin{equation}
    \operatorname{Fix}(\mathrm{NRD})
    =
    \operatorname{Fix}(\mathrm{TB})
    =
    \{(p^\star,Z^\star)\}.
\end{equation}
The usual additive constant in the logits leaves $p$ unchanged.
\subsection{Spurious fixed points of the full gradient and Naive NRD}
\label{app:spurious-fixed-points}
For comparison, consider the on-policy two-trajectory dynamics
\begin{equation}
    \dot\Phi_i
    =
    p_i
    \left[
        -2(\delta_i-\bar\delta)
        +
        s\left(
            \delta_i^2-\overline{\delta^2}
        \right)
    \right],
    \qquad
    \dot{\log Z}
    =
    -2\bar\delta,
    \label{eq:app-symmetric-score}
\end{equation}
where $s=-1$ gives the full expected-loss gradient and $s=+1$ gives Naive NRD.
Take
\begin{equation}
    p_1=\frac14,
    \qquad
    p_2=\frac34.
\end{equation}
For the full gradient, choose
$\delta_1=-3$ and $\delta_2=1$. Then
$\bar\delta=0$ and $\overline{\delta^2}=3$, and
\begin{equation}
    \begin{aligned}
        -2\delta_1-(\delta_1^2-\overline{\delta^2})
        &=6-(9-3)=0,\\
        -2\delta_2-(\delta_2^2-\overline{\delta^2})
        &=-2-(1-3)=0.
    \end{aligned}
\end{equation}
Thus Equation~\ref{eq:app-symmetric-score} is stationary despite the
nonzero residuals.
For Naive NRD, choose instead $\delta_1=3$ and $\delta_2=-1$. Again
$\bar\delta=0$ and $\overline{\delta^2}=3$, while
\begin{equation}
    \begin{aligned}
        -2\delta_1+(\delta_1^2-\overline{\delta^2})
        &=-6+(9-3)=0,\\
        -2\delta_2+(\delta_2^2-\overline{\delta^2})
        &=2+(1-3)=0.
    \end{aligned}
\end{equation}
These residual configurations correspond to valid Trajectory Balance
problems. Given either pair $(p_i,\delta_i)$ above, define
\begin{equation}
    C
    =
    \sum_i p_i e^{-\delta_i},
    \qquad
    p_i^\star
    =
    \frac{p_i e^{-\delta_i}}{C},
    \qquad
    \log\frac{Z}{Z^\star}
    =
    -\log C.
\end{equation}
Then
$\delta_i=\log(Z/Z^\star)+\log(p_i/p_i^\star)$ exactly. Taking two
terminal objects with $P_B(\tau_i\mid x_i)=1$ and rewards proportional to
$p_i^\star$ realizes these examples as valid GFlowNet instances.
\subsection{Proof of Theorem~\ref{thm:pointwise-dominance}}
Let
\begin{equation}
    r_i
    =
    \log\frac{p_i}{p_i^\star},
\end{equation}
and let $i_+$ and $i_-$ attain the maximum and minimum in the definition of
$D$. Since
\begin{equation}
    \delta_i
    =
    \log\frac{Z}{Z^\star}+r_i,
\end{equation}
the same indices maximize and minimize the residual.
The only difference between the NRD and TB logit fields in
Equation~\ref{eq:app-mixture-dynamics} is
\begin{equation}
    (1-\varepsilon)p_i
    \left(
        L_i^--\overline{L^-}_p
    \right).
    \label{eq:app-score-difference}
\end{equation}
The map $x\mapsto x^2\mathbf{1}\!\{x<0\}$ is non-increasing. Hence
\begin{equation}
    L_{i_+}^-
    \leq
    \overline{L^-}_p
    \leq
    L_{i_-}^-.
\end{equation}
The correction in Equation~\ref{eq:app-score-difference} is therefore
non-positive at a maximizing trajectory and non-negative at a minimizing
trajectory.
Since
\begin{equation}
    \dot r_i
    =
    \dot\Phi_i
    -
    \sum_jp_j\dot\Phi_j,
\end{equation}
the common softmax term cancels when differentiating the difference between
the maximum and minimum log-ratios. At every differentiability point,
\begin{equation}
    \dot D_{\mathrm{NRD}}
    -
    \dot D_{\mathrm{TB}}
    =
    (1-\varepsilon)
    \left[
        p_{i_+}
        \left(
            L_{i_+}^--\overline{L^-}_p
        \right)
        -
        p_{i_-}
        \left(
            L_{i_-}^--\overline{L^-}_p
        \right)
    \right]
    \leq0.
\end{equation}
When either extremum is attained by more than one trajectory, the upper
Dini derivative of the maximum is the largest derivative among the active
maximizers, while that of the minimum is the smallest derivative among the
active minimizers. The same sign argument applies to every active index,
which proves the corresponding Dini-derivative inequality.
\subsection{Proof of Corollary~\ref{cor:convergence-strict}}
We first consider the on-policy case. Let
$\bar\delta=\sum_i p_i\delta_i$, and choose $i_+$ and $i_-$ attaining the
maximum and minimum log-ratios. Since the residuals differ from the
log-ratios by the same additive constant,
\begin{equation}
    \delta_{i_+}-\delta_{i_-}
    =
    D.
\end{equation}
For TB,
\begin{equation}
    \dot D_{\mathrm{TB}}
    =
    -2p_{i_+}
    \left(
        \delta_{i_+}-\bar\delta
    \right)
    -
    2p_{i_-}
    \left(
        \bar\delta-\delta_{i_-}
    \right)
    <0
\end{equation}
whenever $D>0$. Theorem~\ref{thm:pointwise-dominance} then gives
$\dot D_{\mathrm{NRD}}\leq\dot D_{\mathrm{TB}}<0$.
To establish convergence, let $D_0=D(0)$. Since $D$ is non-increasing,
all log-ratios
$r_i=\log(p_i/p_i^\star)$ remain inside an interval of width $D_0$.
Writing $m=\min_i r_i$ and using
$\sum_i p_i^\star e^{r_i}=1$ yields
\begin{equation}
    -D_0
    \leq
    m
    \leq
    0.
\end{equation}
Therefore
\begin{equation}
    p_i(t)
    =
    p_i^\star e^{r_i(t)}
    \geq
    c,
    \qquad
    c
    :=
    e^{-D_0}\min_i p_i^\star
    >0.
\end{equation}
The weighted-average identities then give
\begin{equation}
    \begin{aligned}
        \delta_{i_+}-\bar\delta
        &=
        \sum_jp_j(\delta_{i_+}-\delta_j)
        \geq
        p_{i_-}D
        \geq cD,\\
        \bar\delta-\delta_{i_-}
        &=
        \sum_jp_j(\delta_j-\delta_{i_-})
        \geq
        p_{i_+}D
        \geq cD.
    \end{aligned}
\end{equation}
Consequently,
\begin{equation}
    \dot D_{\mathrm{NRD}}
    \leq
    \dot D_{\mathrm{TB}}
    \leq
    -4c^2D,
\end{equation}
and hence
\begin{equation}
    D(t)
    \leq
    D(0)e^{-4c^2t}.
\end{equation}
Thus $D(t)\to0$. Since
$\sum_i p_i^\star e^{r_i(t)}=1$, the common limiting value of the
$r_i$ must be zero, and therefore $p(t)\to p^\star$.
It remains to consider the normalizer. Writing
$\alpha=\log(Z/Z^\star)$,
\begin{equation}
    \bar\delta
    =
    \alpha
    +
    \sum_i p_i
    \log\frac{p_i}{p_i^\star}
    =
    \alpha
    +
    D_{\mathrm{KL}}(p\Vert p^\star),
\end{equation}
and hence
\begin{equation}
    \dot\alpha
    =
    -2\left(
        \alpha+D_{\mathrm{KL}}(p\Vert p^\star)
    \right).
\end{equation}
Since $p(t)\to p^\star$, the KL term converges to zero, and the stable
linear equation above gives $\alpha(t)\to0$. Therefore
$Z(t)\to Z^\star$.
We finally prove the strict comparison under conditional normalization.
Suppose
\begin{equation}
    \bar\delta_q
    =
    \mathbb{E}_{Q_\varepsilon}[\delta]
    =
    0
\end{equation}
and $D>0$. Since $q$ has full support, the largest and smallest residuals
must satisfy
\begin{equation}
    \delta_{i_+}>0>\delta_{i_-}.
\end{equation}
The TB part of Equation~\ref{eq:app-mixture-dynamics} therefore gives
\begin{equation}
    \dot D_{\mathrm{TB}}
    =
    -2
    \left(
        q_{i_+}\delta_{i_+}
        -
        q_{i_-}\delta_{i_-}
    \right)
    <0.
\end{equation}
Moreover, $L_{i_+}^-=0$ while
$\overline{L^-}_p>0$. For every $\varepsilon<1$, the NRD correction at
the maximizing trajectory is therefore strictly negative, while its
correction at the minimizing trajectory is non-negative. Hence
\begin{equation}
    \dot D_{\mathrm{NRD}}
    <
    \dot D_{\mathrm{TB}}
    <0.
\end{equation}
For $\varepsilon=1$, $Q_\varepsilon=\mu$ is independent of the forward
policy and the NRD score term vanishes.
\subsection{Proof of Theorem~\ref{thm:expfam-hitting}}
For the exponential family in
Theorem~\ref{thm:expfam-hitting},
\begin{equation}
    \log
    \frac{p_\eta(\tau)}
    {p_{\eta^\star}(\tau)}
    =
    (\eta-\eta^\star)T(\tau)
    -
    \left(
        A(\eta)-A(\eta^\star)
    \right).
\end{equation}
Under fast normalization,
$\mathbb{E}_{p_\eta}[\delta_\eta]=0$. Subtracting the expectation of the
log-ratio therefore gives
\begin{equation}
    \delta_\eta(\tau)
    =
    (\eta-\eta^\star)
    \left(
        T(\tau)-\mathbb{E}_{p_\eta}[T]
    \right)
    =
    (\eta-\eta^\star)
    \partial_\eta\log p_\eta(\tau).
    \label{eq:app-expfam-residual}
\end{equation}
The TB and NRD parameter dynamics are
\begin{equation}
    \begin{aligned}
        \dot\eta_{\mathrm{TB}}
        &=
        -2
        \mathbb{E}_{p_\eta}
        \left[
            \delta_\eta
            \partial_\eta\log p_\eta
        \right],\\
        \dot\eta_{\mathrm{NRD}}
        &=
        \dot\eta_{\mathrm{TB}}
        +
        \mathbb{E}_{p_\eta}
        \left[
            \delta_\eta^2
            \mathbf{1}\!\{\delta_\eta<0\}
            \partial_\eta\log p_\eta
        \right].
    \end{aligned}
\end{equation}
Multiplying by $\eta-\eta^\star$ and using
Equation~\ref{eq:app-expfam-residual} yields
\begin{equation}
    \begin{aligned}
        (\eta-\eta^\star)\dot\eta_{\mathrm{TB}}
        &=
        -2\mathbb{E}_{p_\eta}
        \left[
            \delta_\eta^2
        \right],\\
        (\eta-\eta^\star)\dot\eta_{\mathrm{NRD}}
        &=
        -2\mathbb{E}_{p_\eta}
        \left[
            \delta_\eta^2
        \right]
        +
        \mathbb{E}_{p_\eta}
        \left[
            \delta_\eta^3
            \mathbf{1}\!\{\delta_\eta<0\}
        \right].
    \end{aligned}
    \label{eq:app-expfam-speed}
\end{equation}
For $\eta\neq\eta^\star$, the sufficient statistic is nonconstant and the
centered random variable in Equation~\ref{eq:app-expfam-residual} takes
both positive and negative values. Hence the last expectation in
Equation~\ref{eq:app-expfam-speed} is strictly negative. This proves
\begin{equation}
    (\eta-\eta^\star)\dot\eta_{\mathrm{NRD}}
    <
    (\eta-\eta^\star)\dot\eta_{\mathrm{TB}}
    <0.
\end{equation}
Both dynamics therefore move monotonically toward $\eta^\star$. A
non-target limiting value would contradict the strict inequality above by
continuity, so both converge to $\eta^\star$.
For a regular one-dimensional exponential family, the Jeffreys divergence is
\begin{equation}
    J(p_\eta,p_{\eta^\star})
    =(\eta-\eta^\star)\bigl(A'(\eta)-A'(\eta^\star)\bigr).
\end{equation}
Indeed, the log-partition terms cancel when the two KL divergences are
added, and $A'(\eta)=\mathbb E_{p_\eta}[T]$. The divergence is finite
for interior natural parameters, including families with unbounded
sufficient statistics. Moreover,
\begin{equation}
    \partial_\eta J
    =A'(\eta)-A'(\eta^\star)
    +(\eta-\eta^\star)A''(\eta).
\end{equation}
Since $A''(\eta)=\operatorname{Var}_{p_\eta}(T)>0$, this derivative
has the sign of $\eta-\eta^\star$. Thus $J$ strictly decreases as the
parameter moves toward $\eta^\star$, and vanishes at the target.
Suppose first that $\eta_0>\eta^\star$. At every common
$\eta>\eta^\star$,
\begin{equation}
    \dot\eta_{\mathrm{NRD}}
    <
    \dot\eta_{\mathrm{TB}}
    <0,
\end{equation}
so NRD has strictly greater speed toward the target. If
$\eta_0<\eta^\star$, division by the negative quantity
$\eta-\eta^\star$ reverses the inequalities and gives
$\dot\eta_{\mathrm{NRD}}>\dot\eta_{\mathrm{TB}}>0$, with the same
interpretation.
Since these are scalar autonomous dynamics, the time required to cross any
interval between the initialization and the target is the integral of the
reciprocal speed over that interval. The strict pointwise speed comparison
therefore yields
\begin{equation}
    T_{\mathrm{NRD}}(\epsilon)
    <
    T_{\mathrm{TB}}(\epsilon)
\end{equation}
for every $0<\epsilon<J(p_{\eta_0},p_{\eta^\star})$: each such threshold
corresponds to a unique parameter between the initialization and the target.
The standard scalar comparison theorem also
gives
$J_{\mathrm{NRD}}(t)<J_{\mathrm{TB}}(t)$ for every $t>0$.
\subsection{Two-trajectory specialization}
\label{app:two-trajectory}
Assume $p>p^\star$; the other case follows after exchanging the trajectory
labels. The discrepancy is
\begin{equation}
    D
    =
    \log
    \frac{p/(1-p)}
    {p^\star/(1-p^\star)}.
\end{equation}
Under fast normalization, the two residuals and the average
negative-residual loss are
\begin{equation}
    \delta_1=(1-p)D,
    \qquad
    \delta_2=-pD,
    \qquad
    \overline{L^-}
    =
    (1-p)p^2D^2.
\end{equation}
For TB, the two free logits therefore satisfy
\begin{equation}
    \dot\Phi_1^{\mathrm{TB}}
    =
    -2p(1-p)D,
    \qquad
    \dot\Phi_2^{\mathrm{TB}}
    =
    2p(1-p)D.
\end{equation}
Since $D=\Phi_1-\Phi_2-\mathrm{const}$,
\begin{equation}
    \dot D_{\mathrm{TB}}
    =
    -4p(1-p)D.
\end{equation}
Only the second trajectory has a negative residual. The NRD score
correction to the two logits is
\begin{equation}
    -p^3(1-p)D^2
    \qquad\text{and}\qquad
    p^3(1-p)D^2,
\end{equation}
respectively. Subtracting the two logit dynamics gives
\begin{equation}
    \dot D_{\mathrm{NRD}}
    =
    -4p(1-p)D
    -
    2p^3(1-p)D^2.
\end{equation}
These equations use two free logits; relative to a single Bernoulli natural
parameter, both fields share the same factor-of-two time rescaling.
The scalar speed comparison therefore gives the trajectory and hitting-time
ordering directly. Here $J=(p-p^\star)D$ is strictly increasing in $D>0$,
so a positive Jeffreys threshold is equivalent to a positive threshold
on $D$ along this branch, consistently with
Theorem~\ref{thm:expfam-hitting}.
It remains to establish the unbounded additive hitting-time advantage.
Along the branch $p>p^\star$,
\begin{equation}
    \frac{p(D)}{1-p(D)}
    =
    \frac{p^\star}{1-p^\star}e^D.
\end{equation}
For a fixed Jeffreys threshold $\epsilon>0$, let $d_\epsilon>0$ be the
unique solution of $(p(d_\epsilon)-p^\star)d_\epsilon=\epsilon$.
The corresponding hitting times, for $D(0)>d_\epsilon$, are
\begin{equation}
    \begin{aligned}
        T_{\mathrm{TB}}(\epsilon)
        &=
        \int_{d_\epsilon}^{D(0)}
        \frac{dD}
        {4p(D)(1-p(D))D},
        \\
        T_{\mathrm{NRD}}(\epsilon)
        &=
        \int_{d_\epsilon}^{D(0)}
        \frac{dD}
        {4p(D)(1-p(D))D
        \left(1+\frac12p(D)^2D\right)}.
    \end{aligned}
\end{equation}
Their difference is
\begin{equation}
    T_{\mathrm{TB}}(\epsilon)
    -
    T_{\mathrm{NRD}}(\epsilon)
    =
    \int_{d_\epsilon}^{D(0)}
    \frac{p(D)}
    {8(1-p(D))
    \left(1+\frac12p(D)^2D\right)}
    \,dD.
\end{equation}
Since
\begin{equation}
    \frac{p(D)}{1-p(D)}
    =
    \frac{p^\star}{1-p^\star}e^D,
\end{equation}
the integrand grows asymptotically as a positive constant times
$e^D/D$. Its integral therefore diverges as $D(0)\to\infty$, proving
\begin{equation}
    T_{\mathrm{TB}}(\epsilon)
    -
    T_{\mathrm{NRD}}(\epsilon)
    \longrightarrow\infty.
\end{equation}

\end{document}